\documentclass[sigconf]{acmart}
\usepackage{algorithm}
\usepackage{algorithmic}
\usepackage{enumitem}
\usepackage{comment}
\setcopyright{none}
\copyrightyear{2024}
\acmYear{2024}
\acmDOI{~}%}
\acmISBN{~}
\acmConference[arXiv Preprint]{}{Research Preview}{2024}%%%
\begin{document}

%%
%% The ''title'' command has an optional parameter,
%% allowing the author to define a ''short title'' to be used in page headers.
\title{FLARE: FP-Less PTQ and Low-ENOB ADC Based AMS-PiM for Error-Resilient, Fast, and Efficient Transformer Acceleration}
%FLARE: FP-Less PTQ and Low-ENOB ADC Based AMS-PiM for Accurate, Fast, and Efficient Transformer Acceleration
%EPL-PiM: Tackling Challenges in Analog-Mixed-Signal PiM Transformer Acceleration with Error Resilience, High Performance, and Low Power Using FP-Less PTQ and Low-ENOB ADC

%%
%% The ''author'' command and its associated commands are used to define
%% ''authornote'' and ''authornotemark'' commands
    
\author{Donghyeon Yi$^1$, Seoyoung Lee$^1$, Jongho Kim$^1$, Junyoung Kim$^1$}
\affiliation{%
  \institution{KAIST$^1$}\country{Republic of Korea}
}
\email{{yicoreen, seoyoung, jongho,lotanda17}@kaist.ac.kr}

\author{Sohmyung Ha$^2$}
\authornote{Corresponding Authors}
\affiliation{%
  \institution{NYU Abu Dhabi$^2$}
  \country{United Arab Emirates}
}
\email{sh169@nyu.edu}

\author{Ik-Joon Chang$^3$}
\authornotemark[1]
\affiliation{%
  \institution{Kyung Hee University$^3$}
  \country{Republic of Korea}
}
\email{ichang@khu.ac.kr}

\author{Minkyu Je$^1$}
\authornotemark[1]
\affiliation{%
  \institution{KAIST$^1$}\country{Republic of Korea}
}
\email{mkje@kaist.ac.kr}

%\affiliation{%
%  \institution{KAIST}
%}
\begin{comment}

\affiliation{%
  \institution{Institute for Clarity in Documentation}
  \city{Dublin}
  \state{Ohio}
  \country{USA}
}

\author{Seoyoung Lee}
\email{seoyoung@kaist.ac.kr}
%\affiliation{%
%  \institution{KAIST}
%}

\author{Jongho Kim}
\email{jongho@kaist.ac.kr}
%\affiliation{%
%  \institution{KAIST}
%}

\author{Junyoung Kim}
\email{lotanda17@kaist.ac.kr}
%\affiliation{%
%  \institution{KAIST}
%}

\author{Sohmyung Ha}
\email{sh169@nyu.edu}
%\affiliation{%
%  \institution{NYU Abu Dhabi}
%}

\author{Ik-Joon Chang}
\email{ichang@khu.ac.kr}
%\affiliation{%
%  \institution{Kyung Hee University}
%}

\author{Minkyu Je}
\email{mkje@kaist.ac.kr}
%\affiliation{%
%  \institution{KAIST}
%}

\end{comment}

%\renewcommand{\shortauthors}{}

\begin{abstract}
    Encoder-based transformers, powered by self-attention layers, have revolutionized machine learning with their context-aware representations. However, their quadratic growth in computational and memory demands presents significant bottlenecks. %for efficient inference in large-scale models. 
    Analog-Mixed-Signal Process-in-Memory (AMS-PiM) architectures address these challenges by enabling efficient on-chip processing.
    Traditionally, AMS-PiM relies on Quantization-Aware Training (QAT), which is hardware-efficient but requires extensive retraining to adapt models to AMS-PiMs, making it increasingly impractical for transformer models. 
    Post-Training Quantization (PTQ) mitigates this training overhead but introduces significant hardware inefficiencies. PTQ relies on dequantization-quantization (DQ-Q) processes, floating-point units (FPUs), and high-ENOB (Effective Number of Bits) analog-to-digital converters (ADCs). Particularly, High-ENOB ADCs scale exponentially in area and energy (2$^{\text{ENOB}}$), reduce sensing margins, and increase susceptibility to process, voltage, and temperature (PVT) variations, further compounding PTQ’s challenges in AMS-PiM systems.
    To overcome these limitations, we propose FLARE, an AMS-PiM architecture that eliminates DQ-Q processes, introduces FPU- and division-free nonlinear processing, and employs a low-ENOB-ADC-based sparse Matrix Vector multiplication technique. Using the proposed techniques, FLARE improves error resiliency, area/energy efficiency, and computational speed while preserving numerical stability. Experimental results demonstrate that FLARE outperforms state-of-the-art GPUs and conventional PiM architectures in energy efficiency, latency, and accuracy, making it a scalable solution for the efficient deployment of transformers.
    %This paper introduces FLARE, an Analog-Mixed-Signal Process-in-Memory (AMS-PiM) architecture designed to efficiently accelerate inference in encoder-based transformer models, which face severe quadratic growth in computational and memory demands due to their self-attention layers. At the core of FLARE architecture is a novel integer-only post-training quantization (PTQ) method that seamlessly integrates with end-to-end self-attention processes, effectively eliminating the need for a high Effective Number of Bits (ENOB) analog-to-digital converters (ADCs) and Floating-Point Units (FPUs). This approach minimizes hardware complexity and energy usage while ensuring accurate and efficient inference. Our FLARE further enhances performance through BitSift-GEMV, a sparse GEneral Matrix-Vector multiplication (spGEMV) technique that exploits bitwise sparsity to reduce latency and increase throughput. Built on a hybrid MRAM-SRAM architecture, FLARE enables on-chip processing of self-attention layers, significantly cutting down tensor traffic and energy consumption compared to traditional PiM and GPU baselines. This architecture provides a scalable, energy-efficient framework for deploying encoder-based transformers in data centers and edge devices, addressing the unique inference-time bottlenecks of these models. 
\end{abstract}

\begin{comment}
    \begin{CCSXML}
<ccs2012>
 <concept>
  <concept_id>00000000.0000000.0000000</concept_id>
  <concept_desc>Do Not Use This Code, Generate the Correct Terms for Your Paper</concept_desc>
  <concept_significance>500</concept_significance>
 </concept>
 <concept>
  <concept_id>00000000.00000000.00000000</concept_id>
  <concept_desc>Do Not Use This Code, Generate the Correct Terms for Your Paper</concept_desc>
  <concept_significance>300</concept_significance>
 </concept>
 <concept>
  <concept_id>00000000.00000000.00000000</concept_id>
  <concept_desc>Do Not Use This Code, Generate the Correct Terms for Your Paper</concept_desc>
  <concept_significance>100</concept_significance>
 </concept>
 <concept>
  <concept_id>00000000.00000000.00000000</concept_id>
  <concept_desc>Do Not Use This Code, Generate the Correct Terms for Your Paper</concept_desc>
  <concept_significance>100</concept_significance>
 </concept>
</ccs2012>
\end{CCSXML}

\ccsdesc[500]{Do Not Use This Code~Generate the Correct Terms for Your Paper}
\ccsdesc[300]{Do Not Use This Code~Generate the Correct Terms for Your Paper}
\ccsdesc{Do Not Use This Code~Generate the Correct Terms for Your Paper}
\ccsdesc[100]{Do Not Use This Code~Generate the Correct Terms for Your Paper}

\end{comment}

%% A ''teaser'' image appears between the author and affiliation
\begin{comment}
\received{20 February 2007}
\received[revised]{12 March 2009}
\received[accepted]{5 June 2009}
\end{comment}
\maketitle
%%%%%%%%%%%%%%%%%%%%%%%%%%%%%%%%%%%%%%%%%%%%%%%%%%%%%%%%%%%%%%%%%%%%%%%%%%%%%%%%%%%%%%%%%%%%%%%%%%%%%%%%%%%%%%%%%%%%%%%%%%%%%%%%%%%%%%%%%%%%%%%%%%%%%%%
%%%%%%%%%%%%%%%%%%%%%%%%%%%%%%%%%%%%%%%%%%%%%%%%%%%%%%%     N E W    S E C T I O N     %%%%%%%%%%%%%%%%%%%%%%%%%%%%%%%%%%%%%%%%%%%%%%%%%%%%%%%%%%%%%%%%
%%%%%%%%%%%%%%%%%%%%%%%%%%%%%%%%%%%%%%%%%%%%%%%%%%%%%%%%%%%%%%%%%%%%%%%%%%%%%%%%%%%%%%%%%%%%%%%%%%%%%%%%%%%%%%%%%%%%%%%%%%%%%%%%%%%%%%%%%%%%%%%%%%%%%%%

\section{Introduction}\label{sec:Intro}
Transformer models~\cite{vaswani2017attention,brown2020language,achiam2023gpt,devlin2018bert,li2023vit,dosovitskiy2020image}, built on the attention mechanism~\cite{vaswani2017attention}, have become foundational in machine learning across diverse domains. 
%Transformer models~\cite{vaswani2017attention,brown2020language,achiam2023gpt,devlin2018bert,li2023vit,dosovitskiy2020image}, built on the attention mechanism~\cite{vaswani2017attention}, have become widely adopted across various domains. 
While decoder-based architectures~\cite{brown2020language,achiam2023gpt}, such as those powering chatbot services, have gained substantial popularity, encoder-based models~\cite{devlin2018bert,li2023vit,dosovitskiy2020image}, like BERT and Vision Transformers (ViT), remain critical for many tasks. 
%While decoder-based architectures~\cite{brown2020language,achiam2023gpt}, such as those used in chatbot services, have gained substantial popularity, encoder-based models~\cite{devlin2018bert,li2023vit,dosovitskiy2020image} like BERT and Vision Transformers (ViT) are also widely embraced. 
However, it is important to note that encoder-based transformers process entire input sequences within their self-attention layers. 
Due to this processing, as the sequence length increases, the computational intensity and intermediate memory traffic of self-attention layers grow quadratically, posing a significant challenge in efficiently processing encoder-based transformers especially in edge devices.
%However, encoder-based transformers process entire input sequences within self-attention layers, leading to computational intensity and intermediate memory traffic that grow quadratically with sequence length. These scaling significantly hinder the efficient deployment of encoder-based transformers, especially in edge devices.

\begin{figure}[t]
    \centering
    \includegraphics[width=0.9\linewidth]{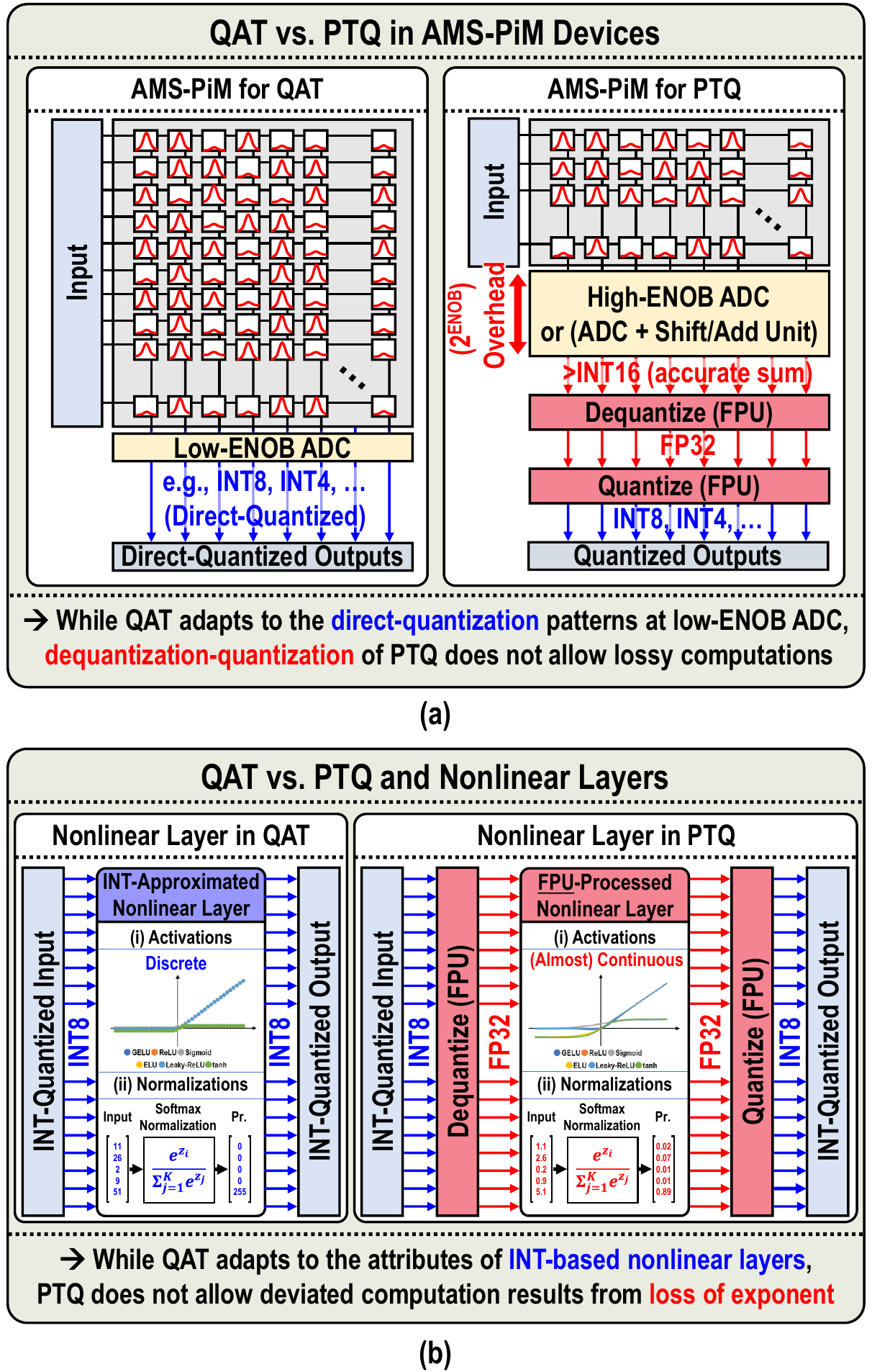}
    \vspace{-2mm} 
    \caption{Limitations of PTQ during inference optimizations. (a) The Dequantization-Quantization (DQ-Q) process and (b) nonlinear layers induce high-ENOB ADCs and FPUs that induce extreme area/energy overhead.}
    \label{fig:introduction}
    \vspace{-4mm}
\end{figure}

Analog-Mixed-Signal Process-in-Memory (AMS-PiM)~\cite{shafiee2016isaac,chi2016prime,jung2022crossbar,xue201924,jiang2020c3sram,sridharan2023x,zhou2022transpim,yang2020retransformer,si2019twin,kim2019nand,biswas2018conv,yin2020xnor,roy2021pim,kazemi2021mimhd,lee20223d,8402126} has emerged as a promising solution to address those challenges, enabling efficient execution of level-2 and level-3 BLAS operations (Basic Linear Algebra Subprograms, primarily matrix multiplications) directly within memory. 
These BLAS operations represent a significant portion of the self-attention layers, making AMS-PiM highly effective for reducing off-chip data movement and energy consumption. 
Besides, AMS-PiM’s effectiveness does not extend to non-BLAS tasks, such as quantization processes or nonlinear-layer computations, requiring additional hardware and computational support. 
Concurrently, most AMS-PiM architectures rely on integer quantization techniques for both stationary weights (e.g., weights for Q (query), K (key), V (value), and O projections) and dynamic activations (e.g., input embeddings, Q/K/V values, attention scores, and final outputs), which are crucial for attention operations in transformers.  
Despite recent advancements in low-precision floating-point (FP) quantization methods~\cite{dettmers2024qlora,xiao2023smoothquant}, AMS-PiM remains constrained to integer quantization due to the substantial overhead required for exponent alignment in FP arithmetic, or the significant accuracy degradation if alignment is omitted. 

Quantization-Aware Training (QAT)~\cite{jacob2018quantization,hubara2018quantized,zhou2016dorefa} and Post-Training Quantization (PTQ)~\cite{xiao2023smoothquant,banner2019post} are two primary techniques for enabling quantization. 
%Meanwhile, there are two major types of quantization technique: Quantization-Aware Training (QAT)~\cite{jacob2018quantization,hubara2018quantized,wang2019haq,zhou2016dorefa} and Post-Training Quantization (PTQ)~\cite{xiao2023smoothquant,banner2019post}. 
QAT, widely adopted in AMS-PiM, optimizes area/energy efficiency by enabling direct quantization with low-ENOB (Effective Number of Bits) ADCs (Analog-to-Digital Converters). Through retraining, QAT allows models to adapt to characteristics of ADCs and integer-approximated nonlinear layers, eliminating the need for dequantization-quantization (DQ-Q) steps and FPUs, as visualized in the left side of Fig.~\ref{fig:introduction}-(a). 
%Typically, QAT has been primarily used in AMS-PiM architectures to further optimize area/energy efficiency with small overhead in peripheral hardware, as visualized in the left side of Fig.~\ref{fig:introduction}-(a). 
%QAT achieves area and energy optimization through a retraining process, which allows models to adapt to the nonlinearities of Analog-to-Digital Converters (ADCs) and enables direct quantization using a low Effective Number of Bits (ENOB) ADCs without the need for Dequantization-Quantization (DQ-Q) steps. This direct quantization within low-ENOB ADCs reduces the complexity of peripheral circuits required for AMS-PiM functionality. 
However, QAT’s retraining process introduces substantial overhead, making it increasingly impractical for growing transformer models.
%However, the retraining process in QAT introduces substantial training overhead, which becomes increasingly impractical as transformer models continue to grow in scale.

Accordingly, PTQ has become a preferred method as a retraining-free alternative to Quantization-Aware Training (QAT). However, as shown on the right side of Fig.~\ref{fig:introduction}-(a), PTQ introduces three major obstacles in AMS-PiM architectures as well, including (1) reliance on high-ENOB ADCs, (2) susceptibility to process, voltage, and temperature (PVT) variations, and (3) the need for Floating-Point Units (FPUs).
%Recently, PTQ has become a preferred method to address the training overhead of QAT due to its ability to quantize models without retraining. However, as shown on the right side of Fig.~\ref{fig:introduction}-(a), PTQ introduces its own challenges in AMS-PiM architectures. Specifically, the DQ-Q process presents three major obstacles: (1) the requirement for high-ENOB ADCs, (3) increased computation errors due to Process, Voltage, and Temperature (PVT) variations, and (2) the need for Floating-Point Units (FPUs). 

Typically, PTQ requires precise partial sums and scaling factors to maintain accuracy, for transformer models with large embedding depths ($\geq$1k) and high bit-precision requirements ($\geq$4 bits per activation and weight). These demands necessitate ADCs with ENOB of $\geq$18 bits to prevent quantization errors, as PTQ lacks the adaptability of QAT to lower-ENOB ADCs via retraining. 
Attempting PTQ with low-ENOB ADCs is not impossible, but would result in excessive latency bottleneck due to the immense computation cycles from segmenting computations into smaller units.
%These challenges arise since PTQ requires precise partial sums and scaling factors to maintain accuracy without retraining. Transformer models, with their large embedding depth ($\geq$1k) and bit-precision requirement ($\geq$4'b for each input activation and weight), demand a high Effective Number of Bits (ENOB) in ADCs ($\geq$18'b) to accurately capture intermediate values and prevent quantization errors. Unlike QAT, which can retrain model parameters to adapt to lower-precision ADCs, PTQ lacks adaptability and requires high-ENOB ADCs to capture fine-grained partial sums with high fidelity directly. Attempting to implement PTQ with low-ENOB ADCs would introduce substantial latency as enormous computation steps would be needed to acquire the partial sum with the required precision, creating a system bottleneck. 
The main concern with the High-ENOB ADCs is that they exacerbate hardware inefficiencies, Their area and power consumption scale exponentially with 2$^{\text{ENOB}}$~\cite{danial2018breaking}, and the narrower sensing margins that increase with ENOB make the system more vulnerable to errors caused by PVT variations. 
%Additionally, increasing ENOB exponentially raises area and power consumption (scaling with 2$^{\text{ENOB}}$)~\cite{danial2018breaking} and narrows the sensing margin, increasing susceptibility to Process, Voltage, and Temperature (PVT) variations. 
Furthermore, PTQ’s reliance on accurate scaling factors during the DQ-Q process necessitates FPUs to handle division operations and avoid cumulative rounding errors, adding to system complexity.
%Furthermore, PTQ’s need for accurate scaling factors from lacking adaptability (from retraining) necessitates FPUs to avoid cumulative rounding errors and preserve precise operation during division operations. 

%These challenges arise since PTQ requires an accurate partial sum and division process during the quantization process. For this purpose, ADCs in a PTQ-based AMS-PiM architecture need to have an ENOB higher than 16—and often exceeding 32—for transformer models with a large number of parameters, where, increasing the ENOB of ADCs results in area and power consumption that scales with 2$^{\text{ENOB}}$~\cite{danial2018breaking}. 
%Furthermore, high ENOB also reduces the sensing margin between analog values by 1/2$^{ENOB}$, thereby increasing the likelihood of errors in analog computation due to the PVT variation.

%Additionally, the DQ-Q process necessitates FPUs for precise division, introducing further inefficiencies. For this, we can consider two scenarios: including FPUs in an AMS-PiM processing element and implementing FPUs outside the AMS-PiM processing elements. Clearly, compared to QAT-based AMS-PiM architectures that employ fully integer processing, the above two scenarios have significantly lower area and power efficiencies.    

%On-chip implementations of high ENOB ADCs and FPUs significantly increase the chip area and power, while off-chip PTQ processing exacerbates memory bottleneck, impacting latency, power efficiency, and system scalability. 

Moreover, nonlinear layers for PTQs introduce additional challenges, as depicted in Fig.~\ref{fig:introduction}-(b).
While integer-based approximations of nonlinear functions and normalization layers~\cite{kim2021bert,li2023vit} work effectively in QAT-based systems, they often lead to significant computation deviations when applied to PTQ. As a result, PTQ relies on dequantization-quantization (DQ-Q) processes and floating-point units (FPUs) to ensure accurate nonlinear-layer processing. This dependency further increases system complexity and overhead.
%Moreover, nonlinear layers in PTQ-based models require more complex processing mechanisms and hardware configurations, as shown in Fig.~\ref{fig:introduction}-(b). 
%Although integer-based approximations of nonlinear functions and normalization layers~\cite{kim2021bert,li2023vit} can effectively replace the original FP operations in QAT-based systems, such approximations in PTQ-based systems often lead to significant deviations in computation results. Consequently, the DQ-Q process and FPUs become essential for handling nonlinear layers in PTQ, further increasing system complexity and overhead.

To address these challenges, we propose FLARE, a novel AMS-PiM architecture that holistically tackles PTQ’s inefficiencies. 
FLARE eliminates high-ENOB ADC, division arithmetics, and FPUs while introducing dequantization-free PTQ and integer-processed nonlinear layers. Furthermore, by leveraging low-ENOB-ADC-based sparse General Matrix Vector (GEMV) operations with bitwise sparsity, FLARE ensures fast, error-resilient, and efficient computation. With a detailed analysis of end-to-end attention computations throughout our design, we enable on-device end-to-end kernel fusion. 
%Therefore, in this literature, we develop a novel AMS-PiM architecture that tackles the aforementioned challenges. 
%We propose optimization techniques that lower the ENOB of ADC and remove FPU while ensuring variation-free, area/energy-efficient, and fast computation. 
%Keys to our technique are the dequantization-free PTQ technique paired with novel integer-processed nonlinear layers developed through a detailed analysis of end-to-end attention computations and their (kernel) fusion. 
%Furthermore, we overcome the typical constraints of low-ENOB ADCs by implementing sparse General Matrix-Vector (GEMV) operations with bitwise sparsity, facilitated by a hardware-based method that avoids encoding and decoding, yielding substantial latency improvements without sacrificing energy or area efficiency.

~

In summary, FLARE achieves: 
\begin{itemize}
    \item end-to-end on-chip processing of self-attention layers, reducing quadratic off-chip tensor traffic;
    \item INT-only, yet accurate, dequantization-free PTQ and nonlinear-layer processing,  to maintain precision without high-ENOB ADC, division, or FPUs;
    \item fast, accurate, and efficient sparse GEMV operations with 6-$\sigma$ confidence leveraging low-ENOB ADC within MRAM-SRAM hybrid AMS-PiM arrays.
\end{itemize}
These advancements provide a scalable and energy-efficient solution for transformer inference, addressing the distinct inference-time bottlenecks of encoder-based models.

%%%%%%%%%%%%%%%%%%%%%%%%%%%%%%%%%%%%%%%%%%%%%%%%%%%%%%%%%%%%%%%%%%%%%%%%%%%%%%%%%%%%%%%%%%%%%%%%%%%%%%%%%%%%%%%%%%%%%%%%%%%%%%%%%%%%%%%%%%%%%%%%%%%%%%%
%%%%%%%%%%%%%%%%%%%%%%%%%%%%%%%%%%%%%%%%%%%%%%%%%%%%%%%     N E W    S E C T I O N     %%%%%%%%%%%%%%%%%%%%%%%%%%%%%%%%%%%%%%%%%%%%%%%%%%%%%%%%%%%%%%%%
%%%%%%%%%%%%%%%%%%%%%%%%%%%%%%%%%%%%%%%%%%%%%%%%%%%%%%%%%%%%%%%%%%%%%%%%%%%%%%%%%%%%%%%%%%%%%%%%%%%%%%%%%%%%%%%%%%%%%%%%%%%%%%%%%%%%%%%%%%%%%%%%%%%%%%%

\section{Backgrounds}

\subsection{Analog-Mixed-Signal Process-in-Memory (AMS-PiM)}

\begin{figure}[h]
    \centering
    \vspace{-2mm}
    \includegraphics[width=0.82\linewidth]{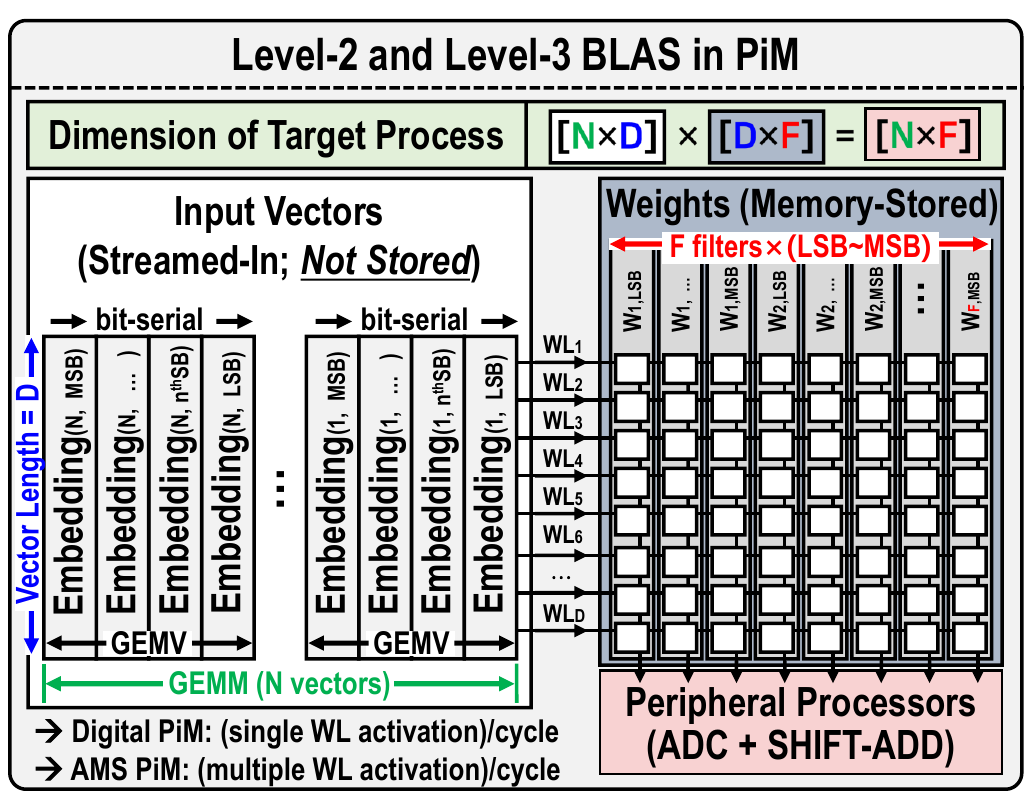}
    \vspace{-2mm}
    \caption{Design example for typical PiM devices, where the level-2 (GEMV) and level-3 (GEMM) BLAS operations are processed inherently.}
    \vspace{-2mm}
    \label{fig:PiM_LinearProjection}
\end{figure}

Process-in-Memory (PiM)~\cite{si201924,chen201865nm,su2017462gops,guo2017fast,zhou2022transpim,jung2022crossbar,verma2019memory,jiang2020c3sram,sridharan2023x,li2024asadi,yang2020retransformer,si2019twin,kim2019nand,yin2020xnor,biswas2018conv,gupta2019nnpim,roy2021pim,kazemi2021mimhd,lee20223d,8402126,imani2019digitalpim,xue201924,sun2018mram} architectures are highly regarded for their ability to execute level-2 (i.e., GEMV) and level-3 (i.e., GEMM) BLAS operations inherently and efficiently, which contribute to a major portion of most deep neural networks (DNNs).
PiM devices allocate matrices onto memory arrays and fetch vectors (which can also form matrices) along wordlines (WLs), bitlines (BLs), or sourcelines (SLs) to execute such BLAS operations, as illustrated in Fig.~\ref{fig:PiM_LinearProjection}. 
In the illustrated configuration, 
weights of $F$ filters are stored across the memory array, 
inputs of depth $D$ are fetched to WL$_{1\sim D}$ bit-serially for m-bit representation, and 
partial sums for each bit position are obtained at BLs.
By integrating BLAS operations~\cite{blackford2002updated} directly within the memory storage, PiM drastically reduces data movement overhead and associated latency. %This enables highly efficient linear projections by storing the weights in the memory matrix and routing input data through wordlines (WLs), bitlines (BLs), and sourcelines (SLs).

Analog-Mixed-Signal Process-in-Memory (AMS-PiM)~\cite{si201924,chen201865nm,su2017462gops,guo2017fast,jung2022crossbar,jiang2020c3sram,sridharan2023x,li2024asadi,
yang2020retransformer,si2019twin,kim2019nand,yin2020xnor,biswas2018conv,lee20223d,8402126} architectures, particularly, enhance the computational efficiency over their digital counterparts. This is achieved by enabling the simultaneous activation of multiple memory interfaces, such as WLs, BLs, and SLs, allowing multiple concurrent computations within a memory array. 
%The simultaneous processing significantly boosts computational throughput and improves efficiency, as illustrated in Fig.~\ref{fig:PiM_LinearProjection}. 
While in digital PiM a single WL is activated per cycle to perform the column sum operations; i.e., pop (``1'' or ``on-cell'')-count operations, similar to a conventional read operation,
contrarily, AMS-PiM simultanesouly activates multiple WLs and converts the analog sum into digital signal via Analog-to-Digital Converters (ADCs), enabling multiple concurrent computations.

%inputs are fetched in parallel to the WLs, then with each column counting the ''1(on)''-cells in the activated WLs within the analog-signal domain. The partial sums generated from these operations are further processed in ALU blocks, enabling multi-bit operand processing to meet the high-precision requirements of modern DNN models.

%\begin{figure}
%    \centering
%    \includegraphics[width=0.9\linewidth]{TensorFootprint_andTypes_Attention.png}
%    \caption{Tensor Footprints along attention computations}
%    \label{fig:AttentionComputation_TensorFootprint}
%\end{figure}
\vspace{-2mm}
\subsection{Self-Attention Layer in Transformer Encoders}
%Transformer-based models, renowned for their efficacy across a myriad of tasks, present unique challenges due to the 
%the attention layers' attributes, which significantly contribute to the complexity and data transfer requirements of these models, as described in Fig.~\ref{fig:introduction}-(b) and (c). 

\begin{figure}[]
    \centering
    \includegraphics[width=\linewidth]{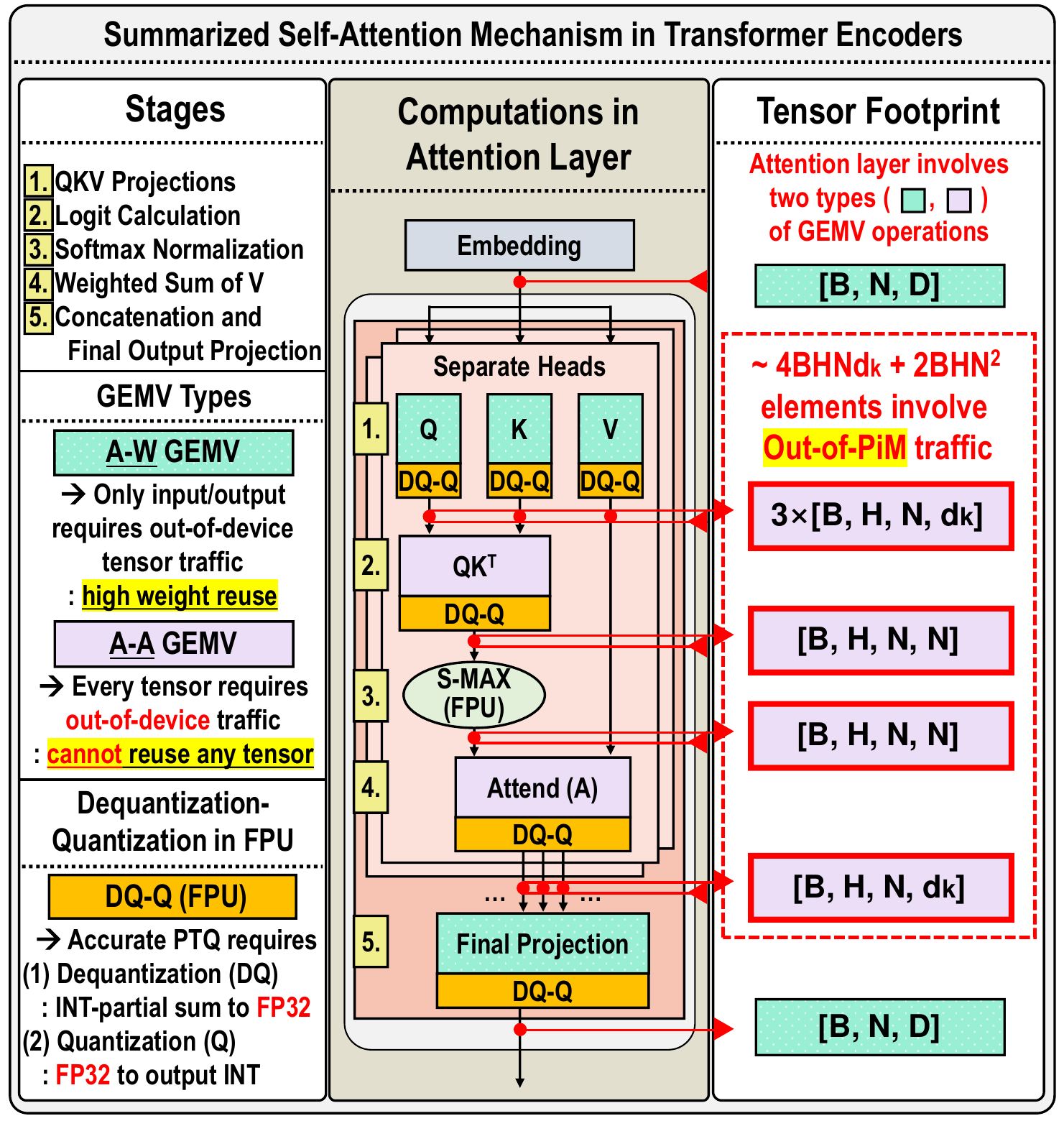}
    \vspace{-5mm}
    \caption{Overview of self-attention, with operation type and tensor traffic analysis. Self-attention comprises multiple linear projection layers with two types of GEMV operations, connected by Softmax, dimensional adjustments (transpose and concatenation), and DQ-Q operations. Unlike QAT, the DQ-Q process and nonlinear layers in PTQ create a deadlock between hardware overhead and tensor traffic.}
    \vspace{-4mm}
    \label{fig:TensorFootprint_Attention}
\end{figure}

For the attention mechanism in a transformer model (described in Fig.~\ref{fig:TensorFootprint_Attention}), each token in the input is represented as a vector of dimension \(D\). Each sequence/image of \(B\) input batches to the self-attention layer consists of \(N\) tokens, resulting in an input tensor of dimension \([B, N, D]\). The input is processed through the attention mechanism with several steps, each contributing to the tensor traffic.

\noindent \textbf{1. QKV Projections:} The initial step involves projecting each input sequence into three distinct weight matrices to produce: Query (Q), Key (K), and Value (V). These projections are computed using stationary weight matrices \(W_Q\), \(W_K\), and \(W_V\) of dimension \([D, d_k]\), where \(d_k\) is a (separate) weight dimension per attention head. 
This results in tensors \(Q = X\cdot W_Q\), \(K = X\cdot W_K\), and \(V = X\cdot W_V\), each with dimension of \([B, N, d_k]\) per head. Given \(H=D/d_k\) heads, the dimension for each Q, K, and V becomes \([B, H, N, d_k]\).

\noindent \textbf{2. Logit Calculation:} The core of the attention mechanism is computing the logit scores \(L\) by performing a dot product between the query and key: \(L = Q\cdot K^T\). This results in a score tensor of dimension \([B, H, N, N]\). The quadratic nature of this operation, requiring \(O(N^2 d)\) operations per head, significantly contributing to the tensor traffic and computational load.

\noindent \textbf{3. Softmax Normalization:} The computed logits are scaled and passed through the Softmax layer to normalize the summed score. 
This normalization step, involving nonlinear operations, 
%, further increases the complexity as it requires \(N\) vector 
%inputfor each token to be processed, maintaining
incorporates tensor traffic of \([B, H, N, N]\).

\noindent \textbf{4. Weighted Sum of Values:} The normalized attention scores are then used to compute a weighted sum of the value (\(V\)) vectors, yielding an output tensor of dimension \([B, H, N, d_k]\).

\noindent \textbf{5. Concatenation and Final Output Projection:} Finally, the outputs from all heads are concatenated and projected together once again, using a weight matrix (\(W_O\)) to produce the final output tensor of \([B, N, D]\).

\noindent \textbf{- DQ-Q Process and FPU:} Not only the Softmax process involving nonlinear functions and accurate divisions, but also each of the stages (from step \textbf{1.} to \textbf{5.}) with PTQ necessitates the DQ-Q process, involving high-ENOB ADC, division, and FPUs. These requirements easily deteriorate the efficacy of PiM, motivating us to design a PiM-based accelerator that alleviates these limitations.    
%Therefore, it is crucial to implement PTQ and nonlinear functions on-chip without any FP or division arithmetic.
%Dequantizes integer values into FP numbers and processes quantization using FPU to return a lower-precision values in a lossless manner. Unlike QAT, the DQ-Q process is a mandatory and burdensome block for PTQ, being crucial maintaining the integrity of the computation process.

\vspace{-2mm}
\subsection{Kernel Fusion and Self-Attention Inference Optimization}

The unique characteristics of the self-attention mechanism have delivered remarkable performance;
however, they also introduce quadratic computational load and memory traffic, 
prompting the development of various innovative techniques to alleviate these limitations.
%These techniques focus on the same goal: reducing 
%tensor footprints and computational complexity. 
Alike PTQ, kernel (operation) fusion techniques also effectively achieve the given goal without impacting the model inference accuracy~\cite{kao2023flat,alwani2016fused,dao2022flashattention}. 
Kernel fusion techniques combine operations involving massive intermediate tensor traffic, reducing memory access overhead for these intermediate tensors.

The out-of-PiM tensor traffic for the self-attention computations without any on-device kernel fusion technique can be expressed as:
\begin{align*}\label{equation:OriginalFootprint}
&\text{Original (Out-of-Device Tensor) traffic} \\ 
&= B \times N \times D \text{ (in)} + B \times N \times D \text{ (out)}\\
&+ 2 \times B \times H \times N^2 + 4 \times B \times H \times N \times d_k \text{ (in-and-out)}.
\end{align*}
As previously mentioned, PiM devices are only suited for processing BLAS layers on-device. Hence, without adequate optimization, the kernel fusion with their inherent limitations - high-ENOB ADCs and FPUs - deteriorate the effectiveness of PiM architectures. 
%with quantization and the nonlinear-function processing involved in the Softmax stage involves overwhelming overhead.

%Furthermore, Unlike conventional hardware like GPUs, PiMs are not innate to those fusion techniques.
%It is improper to process PTQ and nonlinear layers in a conventional manner since it introduces massive overhead, as explained in section~\ref{sec:Intro}.
%hence we must tailor a hardware for core functionality not to overwhelm the main hardware yet not shortfalling in functionality.
Subsequently, our FLARE architecture proposes innovative quantization and nonlinear-layer processing techniques enabling the end-to-end kernel fusion with low hardware overhead, where the revised tensor traffic with our technique is:
\[
\text{Revised Tensor Traffic} = 2 \times B \times N \times D \text{ (in)} + B \times N \times D \text{ (out)} ,
\]
%This reduction is achieved by fusing the operations and eliminating the intermediate footprints, effectively reducing
where the quadratic \(O(N^2)\) tensor traffic reduces to a linear \(O(N)\). 
%For the given example, this translates to a momentous reduction in the tensor footprint, enhancing the efficiency and scalability of the transformer models.

%The techniques include pruning, approximate computing, and tensor compression techniques, significantly cutting down on memory and computational overhead.
%Quantization~\cite{dettmers2022gpt3,xiao2023smoothquant,lin2024awq,kim2021bert,gholami2022survey,shen2020q,lee2024tender,yao2022zeroquant,dettmers2024qlora,frantar2022optq}: implement integer quantization to lower computational complexity and energy consumption, making models more hardware-friendly. Especially, with LLMs, quantization methods that preserves dynamic range for outliers even with integer quantized activation was highlighted and implemented~\cite{lee2024tender,xiao2023smoothquant,dettmers2022gpt3}

%Compression exploiting data pattern~\cite{zaheer2020big,lu2021sanger,ham20203,child2019generating,wang2021spatten}: exploit various types of sparsity, token similarity, locality, etc., along with attention mechanisms to limit the transfer/computation of irrelevant tensors.

%%%%%%%%%%%%%%%%%%%%%%%%%%%%%%%%%%%%%%%%%%%%%%%%%%%%%%%%%%%%%%%%%%%%%%%%%%%%%%%%%%%%%%%%%%%%%%%%%%%%%%%%%%%%%%%%%%%%%%%%%%%%%%%%%%%%%%%%%%%%%%%%%%%%%%%
%%%%%%%%%%%%%%%%%%%%%%%%%%%%%%%%%%%%%%%%%%%%%%%%%%%%%%%     N E W    S E C T I O N     %%%%%%%%%%%%%%%%%%%%%%%%%%%%%%%%%%%%%%%%%%%%%%%%%%%%%%%%%%%%%%%%
%%%%%%%%%%%%%%%%%%%%%%%%%%%%%%%%%%%%%%%%%%%%%%%%%%%%%%%%%%%%%%%%%%%%%%%%%%%%%%%%%%%%%%%%%%%%%%%%%%%%%%%%%%%%%%%%%%%%%%%%%%%%%%%%%%%%%%%%%%%%%%%%%%%%%%%

\section{Motivations}\label{sec:current_limits_of_quantizationtechnique}

%\subsection{Key Points of Our PTQ Optimization}

\begin{figure}[t]
    \centering
    \includegraphics[width=0.9\linewidth]{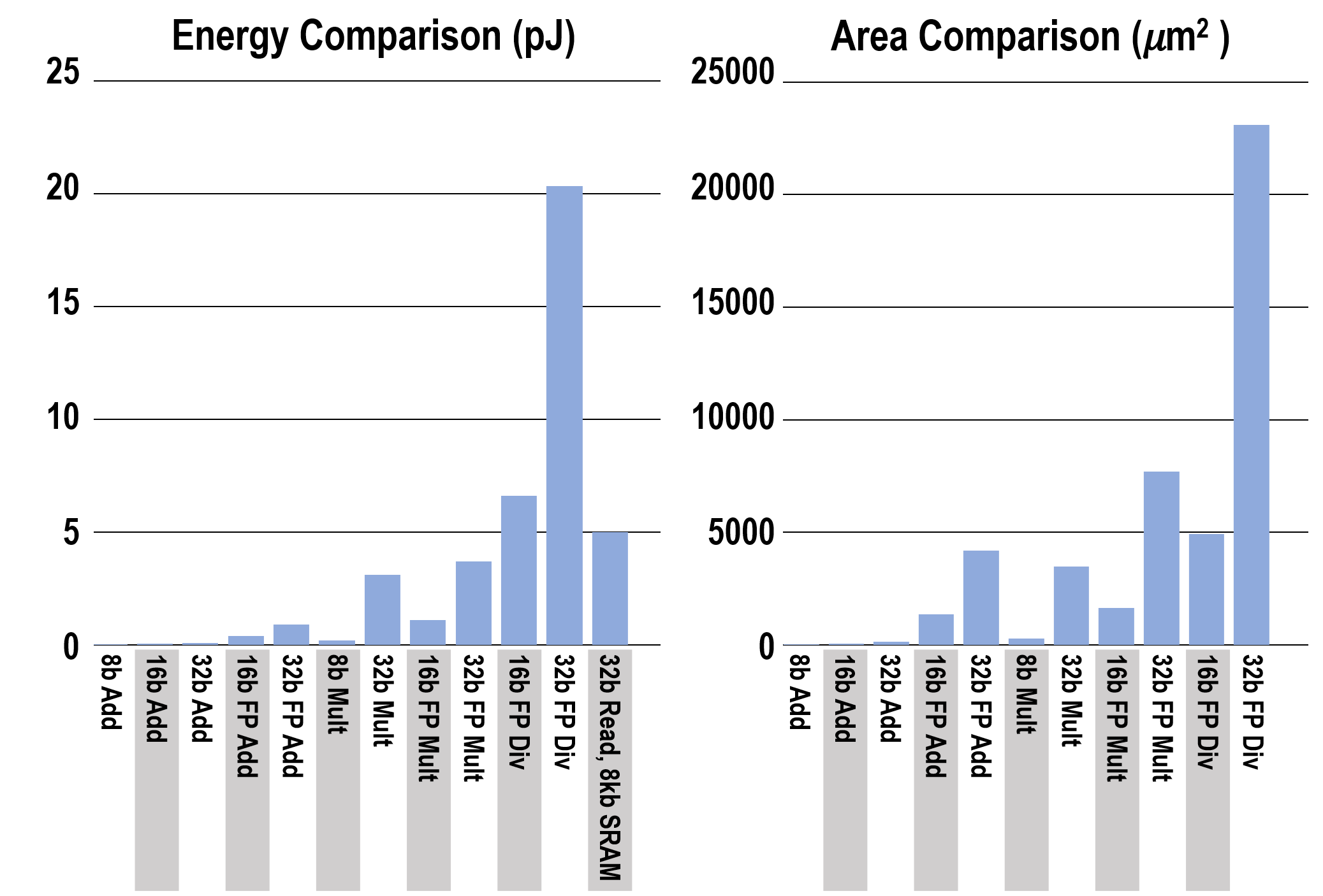}
    \vspace{-2mm}
    \caption{We alleviate not only the reliance on high-ENOB ADCs and FPUs but also the need for division arithmetic in our PiM hardware.
    }
    \label{fig:FP_Division_overhead}
    \vspace{-3mm}
\end{figure}

Our study reveals that the effectiveness of PiM architectures is notably hindered by the high-ENOB ADC and FPUs.
%processing demands required to support PTQ within it. 
%While PiM efficiently and inherently handles integer-quantized BLAS operations, it faces challenges performing DQ-Q process and other nonlinear operations directly on-device. 
%This motivates us to eliminate operations that burden PiM hardware when handling those intricate computations -
Additionally, among various arithmetic operations required for the execution of self-attention, division arithmetic - not only in FP but also in integer formats - hinders the hardware efficiency, as illustrated in Fig.~\ref{fig:FP_Division_overhead}. 
%Thus, we aim to remove those inconvenient arithmetics within our processing method.
%While eliminating such overheads, our optimization strategy aims following ideas:
Thus, our optimization strategy tackles these challenging overheads while preserving accuracy and performance, focusing on the following principles:
\begin{itemize}[leftmargin=*]
    \item \textbf{Accurate Quantization}
    \begin{itemize}[leftmargin=*]
        \item Lossless quantization: Our method must accurately represent quantized values even without FP or division arithmetic.
        \item Outlier preservation: Attention layers often generate channels with extreme outliers, which standard integer quantization in PTQ cannot capture well. To address this, we use token-wise quantization to reflect these outliers accurately.
    \end{itemize} 
    \item \textbf{Accurate Nonlinear (Non-BLAS)-Layer Executions}
    \begin{itemize}
        \item Transformer models depend heavily on nonlinear layers, which require complex exponentiation and division operations. Our approach preserves the exponential trend and proportional relationships between values without FP or division arithmetic.
    \end{itemize}
    \item \textbf{PVT-Robust and Error-Resilient AMS-PiM}
    \begin{itemize}
        \item Despite its high efficiency, AMS-PiM has frequently been criticized for computation errors due to PVT variations. %, where the issue is even more severe with transformer models and PTQ. 
        However, design solutions for those errors often conflict with the efficient exploitation of PiM devices. 
        Our proposed techniques resolve this dilemma effectively.
        %computation method that allows the use of low-ENOB ADCs with a wide sensing margin, while enabling accurate, fast, and area/energy-efficient BLAS operations in AMS-PiM devices.
        %Furthermore, we preserve the dynamic range of channel(word)-wise activation throughout BLAS operation of the end-to-end fusion to preserve the channel-wise outliers of LLMs. 
        %This approach aims to combine the accuracy of FP processing with the efficiency of integer processing, allowing PiM to handle these operations more effectively without imposing a heavy computational burden on the hardware.
    \end{itemize}    
\end{itemize}
In the following sections, we analyze these principles more deeply, before presenting our solutions as a breakthrough.
%In the following sections, we analyze the characteristics of each requirement before presenting our solutions as a breakthrough.

\subsection{Motivation for PTQ and Non-BLAS Layer Optimizations}\label{sec:PTQ_nonBLAS_Optimization}

\begin{figure}[h]
    \centering
    \includegraphics[width=\linewidth]{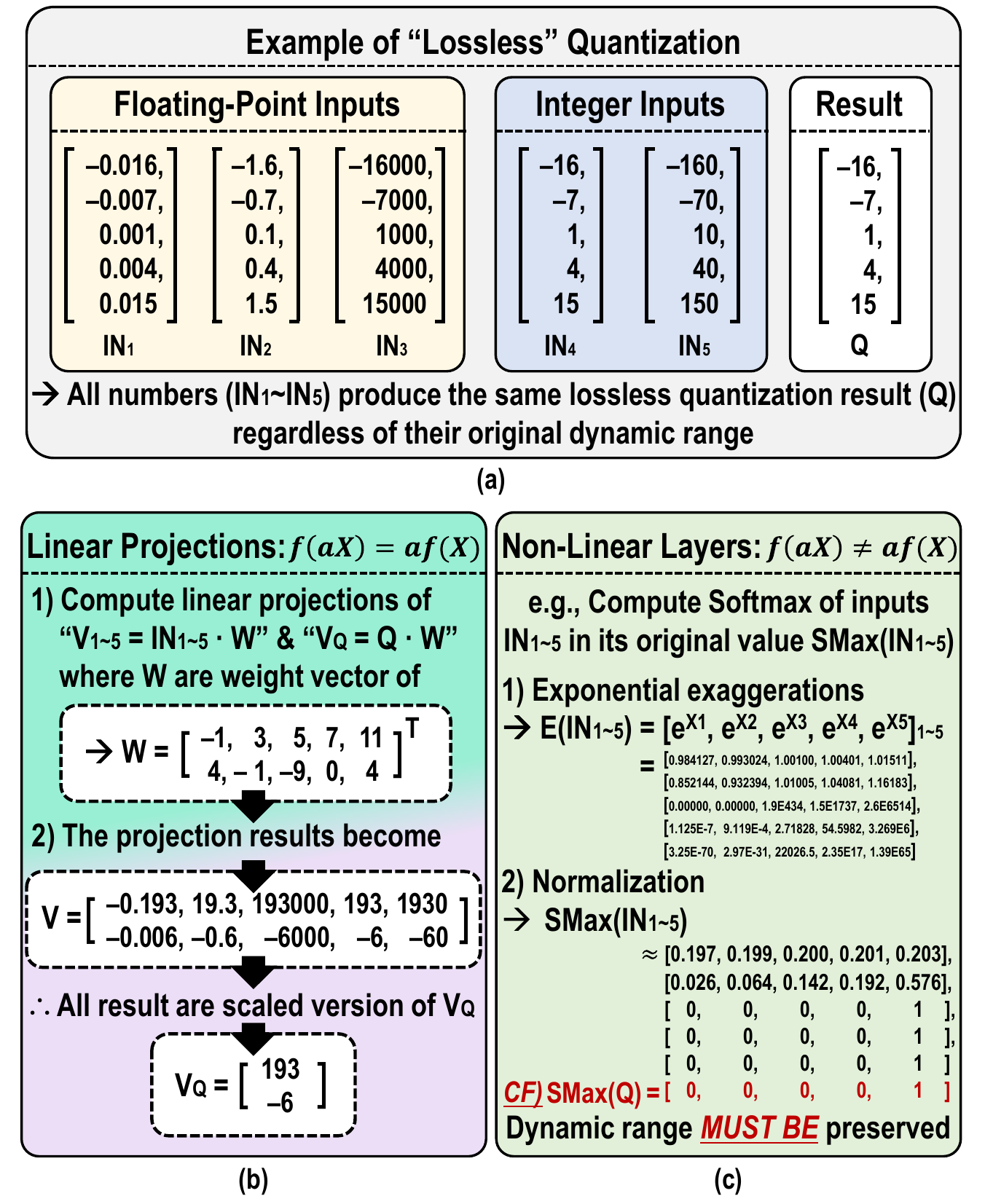}
    \vspace{-7.5mm}
    \caption{(a) Even for a lossless quantization, values lose the exponent information. 
    (b) The linear projections are consistent without exponent. % when we provide sufficient bit-precision at the output. 
    (c) Softmax with the values shown at (a) - the absence of exponents deteriorates the computation result of nonlinear layers.  
    %indicating that exponent-excluded integer-quantized values can deteriorate everything. 
    %The example assumes the symmetric input-output quantization for an easier understanding.
    }
    \vspace{-4mm}
    \label{fig:CurrentLimit_of_Quantization}
\end{figure}

%Our motivation for optimizing PTQ comes from the limitations of integer quantization and the relationship between integer-quantized values and nonlinear-function layers.
While there are numerous quantization methods that we may not be able to cover in this literature, the fundamental steps %that require FP and division arithmetic along DQ-Q
for the accurate yet efficient quantization processes are as follows: 
(1) identify the minimum and maximum values among the given inputs ($x_i\in\mathbf{X}$), 
(2) for n-bit quantization, define a ``quantization step ($\mathbf{S}$),'' by dividing the minimum-maximum range by \(2^n\) levels, and
(3) divide $x_i$'s by $\mathbf{S}$ to obtain the quantized values. 
%During the process, FP division arithmetic is often considered crucial for ensuring accurate division by the quantization step. 

Along the process, as illustrated in Fig.~\ref{fig:CurrentLimit_of_Quantization}-(a), we say a quantization is ``lossless'' when the ratios between quantized values retain the original proportions. That is:
\begin{equation}\label{eq:LosslessQuantization}
    \text{quantize}(x_i)/\text{quantize}(x_j) \approx x_i / x_j,~ \forall x_i, x_j\in\mathbf{X} ~.
\end{equation}
However, the original exponent is not retained, even when ``lossless'' quantization is achieved using FPUs. 
% even with lossless quantization. %as designated. %, while it actually is not. 
Nonetheless, the issue of omitting exponents is less severe in linear projection operations (as depicted in Fig.~\ref{fig:CurrentLimit_of_Quantization}-(b)), as operations keep their inherent quality without exponent information, i.e., 
\begin{equation}
    a\times f(\mathbf{X}) = f(a\times\mathbf{X}).
\end{equation}
On the other hand, nonlinear layers are completely disrupted when the exponent information is omitted, as shown in Fig.~\ref{fig:CurrentLimit_of_Quantization}-(c). That is, 
\begin{equation}
    a\times f(\mathbf{X}) \neq f(a\times\mathbf{X}).
\end{equation} 
Some recent studies simplify FP-based computations~\cite{cardarilli2021pseudo,kim2021bert,li2023vit} to integer-based lightweight alternatives with satisfactory reliability. However, this approach can reduce the accuracy of PTQ-based inference, from the lack of retraining to adapt the model for these approximations.

In response, in section~\ref{sec:fpless_Quantization} and ~\ref{sec:VDRsoftmax}, 
we propose an integrated quantization-computation method that preserves the exponent for nonlinear layers while replacing division operations only with parsing and bit-shifting. While prior works~\cite{ho20232a,anonymous2024qrazor} introduced capturing the MSB’s position, utilizing parsing(s), they rely on group-wise quantization, incurring large storage overhead and limited scalability. In contrast, our method combines token-wise quantization with VDR-Softmax in a fused manner, temporarily storing only a single-token value in compact registers and discarding them after processing, and linking quantization to penetrate through the end-to-end attributes of each stages in attention layer.
%we propose an integrated quantization-computation method that keeps the exponent for nonlinear layers while replacing division operations with parsing and bit-shifting. Additionally, while the idea of capturing the MSB's position is often introduced in some literatures~\cite{ho20232a,anonymous2024qrazor}, our method differs by leveraging token-wise quantization that is utilized with our proposed VDR-Softmax in a fused manner, thus alleviating the large storage for exponents and grouping algorithm which limits scalability,

\subsection{Motivation for Accurate and Efficient AMS Computation}\label{sec:UnreliableComputation}

\begin{figure}[h]
    \centering
    \vspace{-3mm}
    \includegraphics[width=1\linewidth]{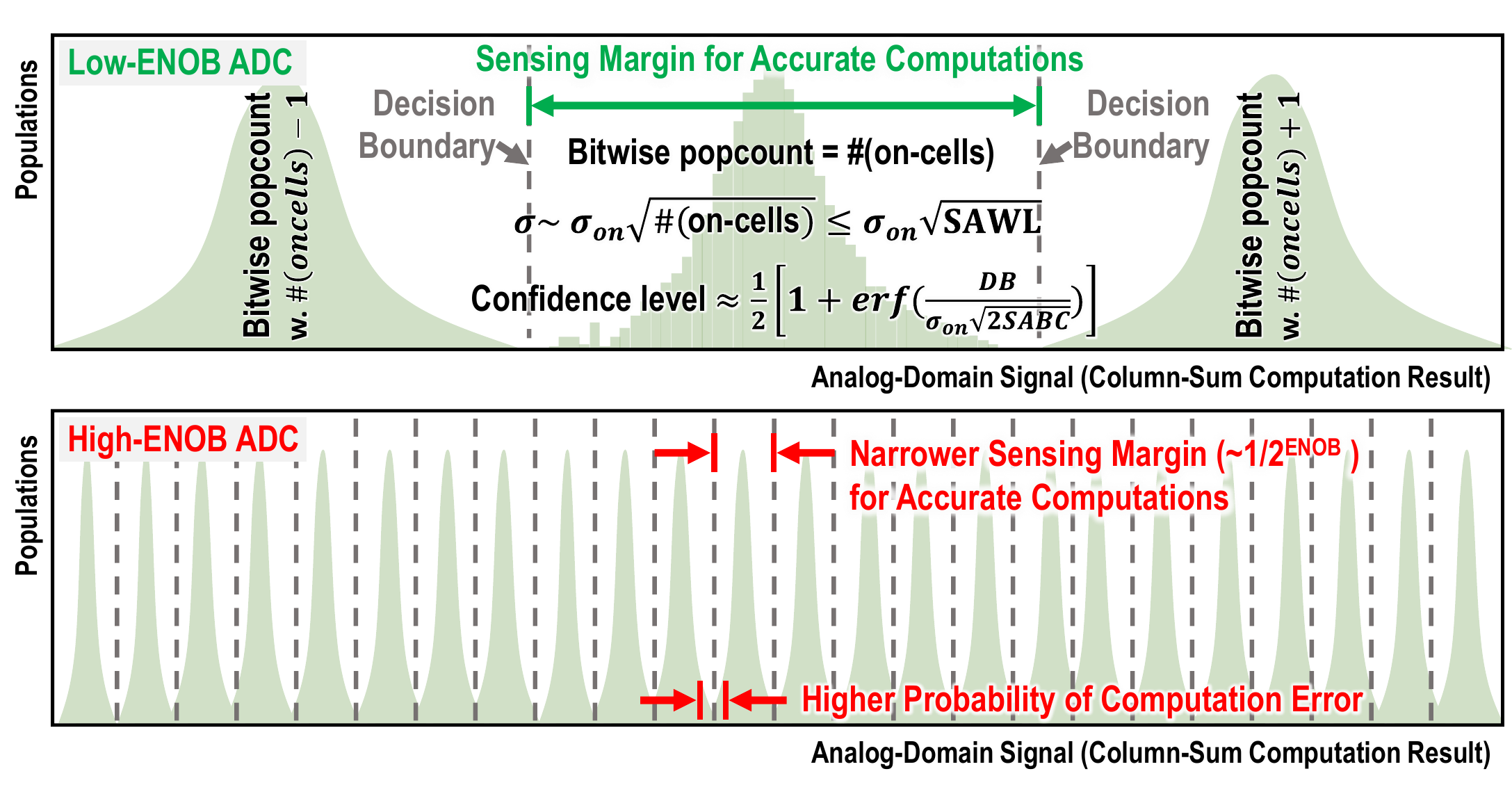}
    \vspace{-5mm}
    \caption{%The greater number of ``on-cells'' involved in a unit computation deteriorates the computation results. The computation's standard deviation and the confidence level for non-overlapping results are derived. $\sigma_{on},$ denotes the standard deviation of ``on-cell'' current, which is much larger than that of the ``off-cell'' ($\because\sigma_{on}\gg\sigma_{off}$). Note that, the standard deviation of the analog sum = ($\sigma_{on}\times\sqrt{\text{number of ``on-cell''s}}$) is upper-bounded by $\sigma_{on}\times\sqrt{\text{SAWL}}$. 
    AMS-PiM with high-ENOB ADCs are more susceptible to computation errors due to PVT variation, resulting from reduced sensing margins.
    }
    \vspace{-2mm}
    \label{fig:ComputeConfidence_SAWL}
\end{figure} 

As mentioned in section~\ref{sec:Intro}, 
transformer models with PTQ demand a high precision for partial sum, easily exceeding $\geq$18 bits with $d_{model}$$\geq$1k and with activation/weight operands of $\geq$4bit. 
This high-precision summation, for which direct quantization is infeasible, requires ADCs with higher ENOBs: this increases not only area/energy overhead but also the likelihood of errors.

Ensuring precise analog computation for PTQ requires the number of decision boundaries to exceed the distinct analog signal levels. This requirement is fundamental in PTQ, as accurate quantization relies on precise summation and digital conversion of the analog values. 
However, achieving this narrows the sensing margin of ADCs, which worsens error resilience, as depicted in Fig.~\ref{fig:ComputeConfidence_SAWL}. Errors occur when analog signals cross decision boundaries during analog-to-digital conversion, and this challenge intensifies as sensing margins (the distance between decision boundaries) narrow.

%The computational error in AMS-PiM can be characterized as illustrated in Fig.~\ref{fig:ComputeConfidence_SAWL}. The computation error occurs when an analog signal intrudes over the decision boundary during the analog-to-digital conversions, and error resilience decreases as the sensing margin (the distance between decision boundaries) narrows.
%The probability of the computational error increases with an increasing ENOB, as the decision boundaries get closer. 
%Ensuring precise analog computation, such as summation in PTQ, requires the number of decision boundaries to exceed the distinct analog signal levels. This fundamental requirement for PTQ directly impacts the sensing margin of ADCs, as the increased precision demands narrow the margin and challenge error resilience. 

The computational error resilience, which prompts us to our proposed technique, can be analyzed as follows. Analog signals are proportional to the number of ``on-cell''s generating on-current within the activated word lines (WLs), as illustrated in Fig.~\ref{fig:PiM_LinearProjection}. 
Recent studies~\cite{yi2024skew,andrulis2023raella,kim2020energy} show that reducing the activated ``on-cell''s along the partial sum process reduces the error rates.
Meanwhile, the number of Simultaneously Activated WLs (SAWL) determines the direct upper limit of the ``on-cell''s contributing to the analog summation. 
Therefore, we can also limit the SAWL to limit the number of active ``on-cell''s and segment unit computations with shorter input vectors for more reliable computation. 
%increasing SAWL narrows the sensing margin, deducting the confidence level of the analog computations being correct, as shown in Fig.~\ref{fig:ComputeConfidence_SAWL}.
This approach reduces the required ENOB of ADC, lowering the area/energy overhead of ADCs by \(2^\text{ENOB}\), significantly relaxing the hardware complexity~\cite{danial2018breaking}.
By integrating strategic segmentation reducing ADC overhead with error mitigation techniques described in section~\ref{Sec:BitSift}, FLARE can significantly improve PVT robustness.

\begin{figure}[]
    \centering
    \includegraphics[width=0.96\linewidth]{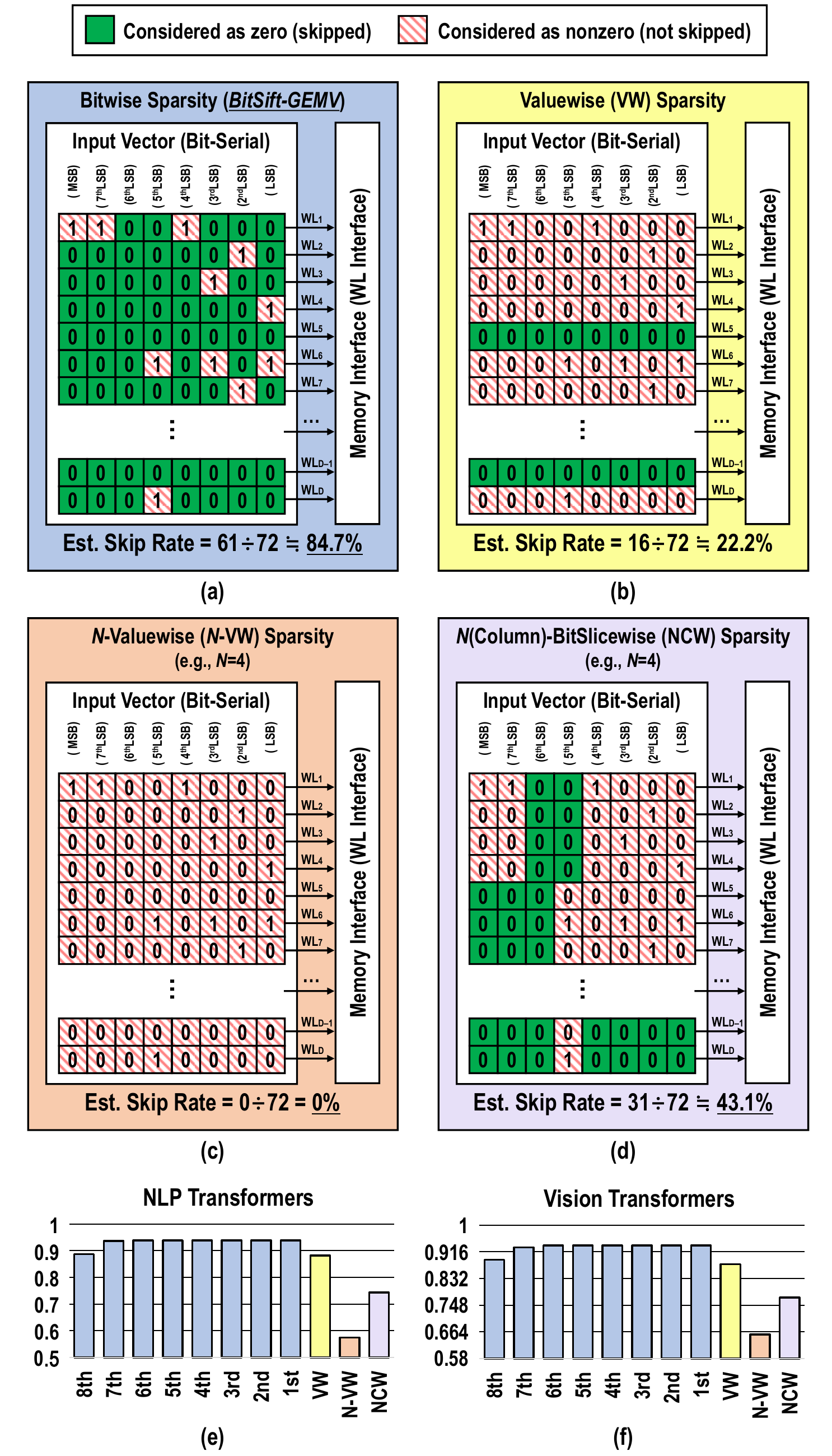}
    \vspace{-3mm}
    \caption{Comparison of sparsity measured in various granularity. 
    %The definition with example with 
    (a)$\sim$(d) visualizes numerous sparsities where the same 8-bit-integer vectors are bit-serially inputted. Sparsities are measured and exploited; (a) bitwise manner, (b) valuewise (VW) manner (when 8-bit integer is of complete ``0''s), \textit{N}-valuewise (\textit{N}-VW) manner, and (d) \textit{N}-column-BitSlicewise manner (NCW, when a part of bitwise input is all ``0''s in column direction). The activation sparsities are measured and averaged with input, Q, K, V, and output layers in transformer blocks for (e) NLP task and (f) vision task.}
    \vspace{-5mm}
    \label{fig:comparison_various_sparsity}
\end{figure}

\subsection{Massive GEMVs with Quadratic Latency Bottleneck}\label{sec:GEMV_quad_latency}

As delineated in Section~\ref{sec:UnreliableComputation}, the error-free analog computation can be achieved by limiting the maximum number of SAWL and segmenting computations over multiple cycles with sliced inputs (e.g., dividing $D$ in Fig.~\ref{fig:PiM_LinearProjection} into $D/N$ using $N$ cycles). 
However, this sliced processing — with a constrained vector length — results in significantly more processing cycles.
Furthermore, the increasing model size and the resultant quadratic GEMV demand in the attention layer further exacerbate the suboptimal performance.

For example, consider a length($l$)-restricted GEMM operation where SAWL$\leq$$l$=8, with: \(D=1024\) hidden states and \(N=512\) tokens per batch - resulting in 512 (=$N$) GEMVs -, using 8-bit integer representation (Bit-Precision, $BP$=8) for both input and weight values. 
This necessitates \(D/l=1024/8=128\) cycles to process a GEMV for each bit-serial vector of a token. 
It involves partitioning each bit-serial part of the input token into 8-WL chunks to limit SAWL$\leq$8.
%to partition each of input embeddings into 8-WL chunks and limit SAWL under 8. 
Subsequently, the bitwise GEMV operations of \(D/l\)(=128)-cycles-per-bit are required for \(N \times BP = 512 \times 8 = 4,096\) times to process every GEMVs of bit-vectors and tokens composing the aforementioned GEMM, culminating in \((D/l) \times N \times BP = 524,288\) cycles to process a GEMM of an embedding input with static weight matrices. 

Assuming 10ns per bitwise GEMV, which is a fast figure for an AMS-PiM array, a GEMM operation for generating Q, K, or V would take approximately 0.5 milliseconds (ms). 
Other steps involving activation-activation GEMV with quadratic computational complexity and nonlinear layers may induce substantially greater latency, rendering the PiM device noncompetitive due to poor latency performance. 
Consequently, the latency for processing only the linear projections across the attention layers in tens of encoder blocks could be approximately in the order of 100 ms.

To address the latency issue, we take advantage of the high ``bitwise'' sparsity found in the activation values of bitwise GEMVs in the attention layer. 
%Numerous existing works utilize sparse data by either compressing tensor data before fetching it into an accelerator or employing zero-skipping techniques at various granularities to enhance energy efficiency and computational latency. 
Our findings suggest collecting only ``1''s along a bitwise embedding input can significantly boost GEMV - and GEMMs composed of GEMVs - operations. 
Our investigation visualized in Fig.~\ref{fig:comparison_various_sparsity}-(e), (f) shows that when split into bitwise elements (Fig.~\ref{fig:comparison_various_sparsity}-(a)), the activation matrices exhibit much higher sparsity than when measured in coarser value grains (Fig.~\ref{fig:comparison_various_sparsity}-(b) $\sim$ (d)).
With our findings, we propose BitSift-GEMV technique, where all bitwise ``0''s of bit-serial fetched input activations are skipped from GEMV within AMS-PiM arrays. 

In section~\ref{Sec:BitSift}, we propose a hardware-oriented technique that designates the longest combination(s) of parsed input slices, incorporating a fixed, desired number of ``1''s. 
By parsing and merging the bit-serial-fetched token vectors, we can significantly stretch the length of a processible input vector for a unit GEMV. 
Furthermore, our proposed technique fully exploits the highest utilization of the designed ADCs, stabilizing its operation by limiting flexibility.  
%Furthermore, compression techniques often involve approximate computing, leading to potential accuracy degradation and limited compatibility with other techniques. Furthermore, conventional hardware configurations constrain the granularity of sparsity processing, typically to a word-level or set-level~\cite{a,b,c,d}. Our proposed BitSift-GEMV in the following section proposes a hardware-oriented

%

%%%%%%%%%%%%%%%%%%%%%%%%%%%%%%%%%%%%%%%%%%%%%%%%%%%%%%%%%%%%%%%%%%%%%%%%%%%%%%%%%%%%%%%%%%%%%%%%%%%%%%%%%%%%%%%%%%%%%%%%%%%%%%%%%%%%%%%%%%%%%%%%%%%%%%%
%%%%%%%%%%%%%%%%%%%%%%%%%%%%%%%%%%%%%%%%%%%%%%%%%%%%%%%     N E W    S E C T I O N     %%%%%%%%%%%%%%%%%%%%%%%%%%%%%%%%%%%%%%%%%%%%%%%%%%%%%%%%%%%%%%%%
%%%%%%%%%%%%%%%%%%%%%%%%%%%%%%%%%%%%%%%%%%%%%%%%%%%%%%%%%%%%%%%%%%%%%%%%%%%%%%%%%%%%%%%%%%%%%%%%%%%%%%%%%%%%%%%%%%%%%%%%%%%%%%%%%%%%%%%%%%%%%%%%%%%%%%%

%\begin{figure*}[ht]
%    \centering
%    \includegraphics[width=\textwidth]{Het_PiM_PE.png}
%    %\vspace{-3mm}
%    \caption{FLARE PE (Processing Element) for an E2E(End-to-End) attention acceleration based on two-type AMs-PiM devices, Fuses batch-input stage to attended-output in E2E manner}
%    \label{fig:AttenPiMPE}
%\end{figure*}

\section{FLARE Architecture}

\begin{figure}[h]
    \centering
    \includegraphics[width=1.01\linewidth]{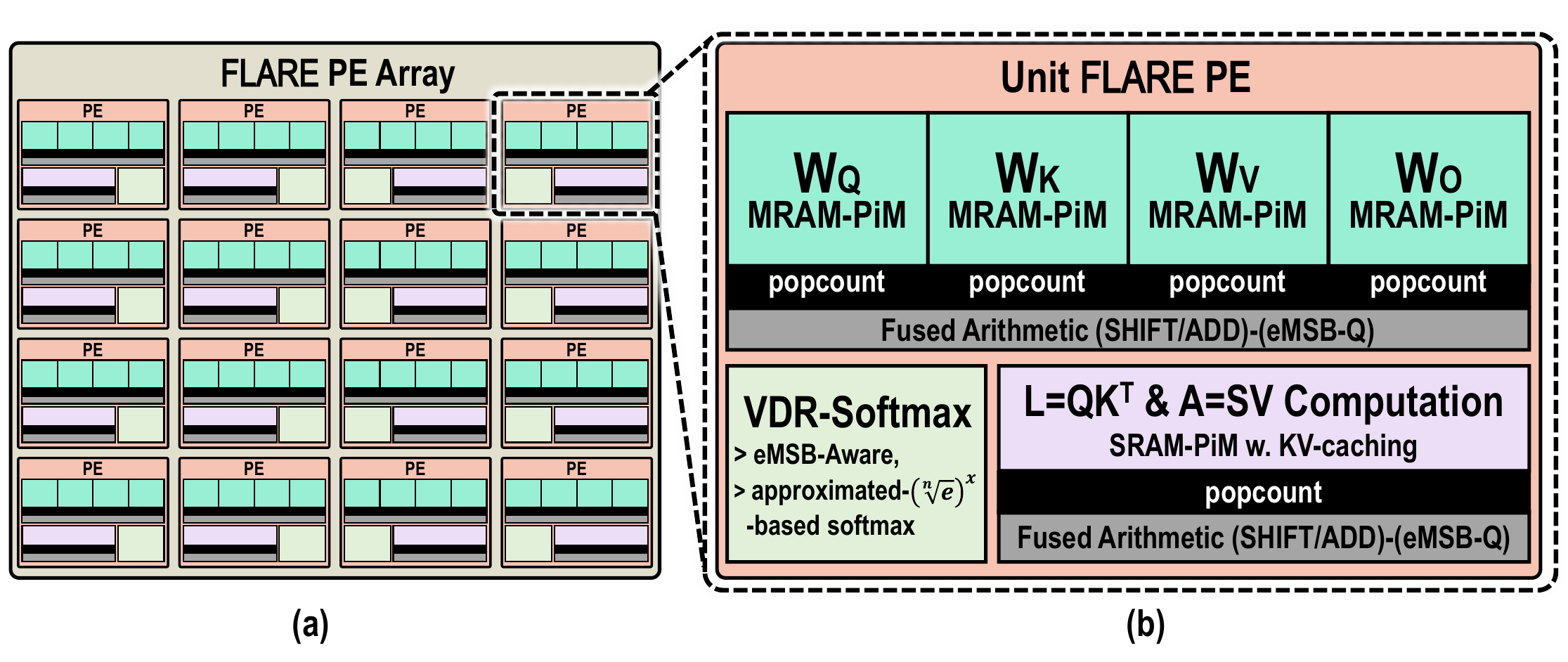}
    \vspace{-7mm}
    \caption{(a) Overall FLARE Architecture. (b) Hardware configurations of each FLARE PE. %based on two-type AMs-PiM devices, fusing from the input stage to the final-projection stage in E2E manner. 
    %The method is assisted by MRAM-SRAM hybrid architecture for A-W/A-A GEMV with fast and lossless BitSift-GEMV technique and efficient fusion of layer connections by proposed FP-less quantization and variable-dynamic-range (VDR) Softmax functions.
        }
    \label{fig:AttenPiMArch}
    \vspace{-3mm}
\end{figure}

Building upon our findings and discussions, we present our FLARE architecture — an FP- and division-less, end-to-end attention accelerator based on MRAM-SRAM hybrid AMS-PiM with low-ENOB ADCs. The architecture overview is visualized in Fig.~\ref{fig:AttenPiMArch}. 
%The proposed architecture suggests an energy-efficient, accurate, and fast attention-layer acceleration method based on an MRAM-SRAM hybrid AMS-PiM arrays. 
Our proposed architecture ensures accurate, reliable, efficient, and fast on-device attention-layer computations and is validated at the post-layout level using a 28nm FD-SOI process. 

%Upon the discussed baseline in the previous chapter, we design an end-to-end attention accelerator architecture. We implemented our hardware using 8-bit integer weights, and variable-bit-precision activations while the quantization of intermediate values is fused into the proposed hardware. Also, we propose a novel, hardware-oriented Softmax processing technique, where integer-friendly operators mimic most of the FP-like operations and contributions. The following chapters discuss how we carefully designed MRAM-SRAM hybrid AMS-PiM devices, GEMV-processing technique, and nonlinear operations to embed end-to-end operations in a compact-sized ASIC with superior Joule/token and latency/token performance. Also, we verify our design physical-level by doing meticulous circuit simulations including Monte-Carlo simulation for mismatch testing, and post-layout simulation for physical-level feasibility, upon our custom designed ASIC using 28nm FD-SOI technology, which supports embedded STT-MRAM(Spin Transfer Torque Magnetic Random Access Memory).

\subsection{MRAM-SRAM Hybrid AMS-PiM Design}

\begin{figure}[h!]
    \centering
    \vspace{-2mm}
    \includegraphics[width=\linewidth]{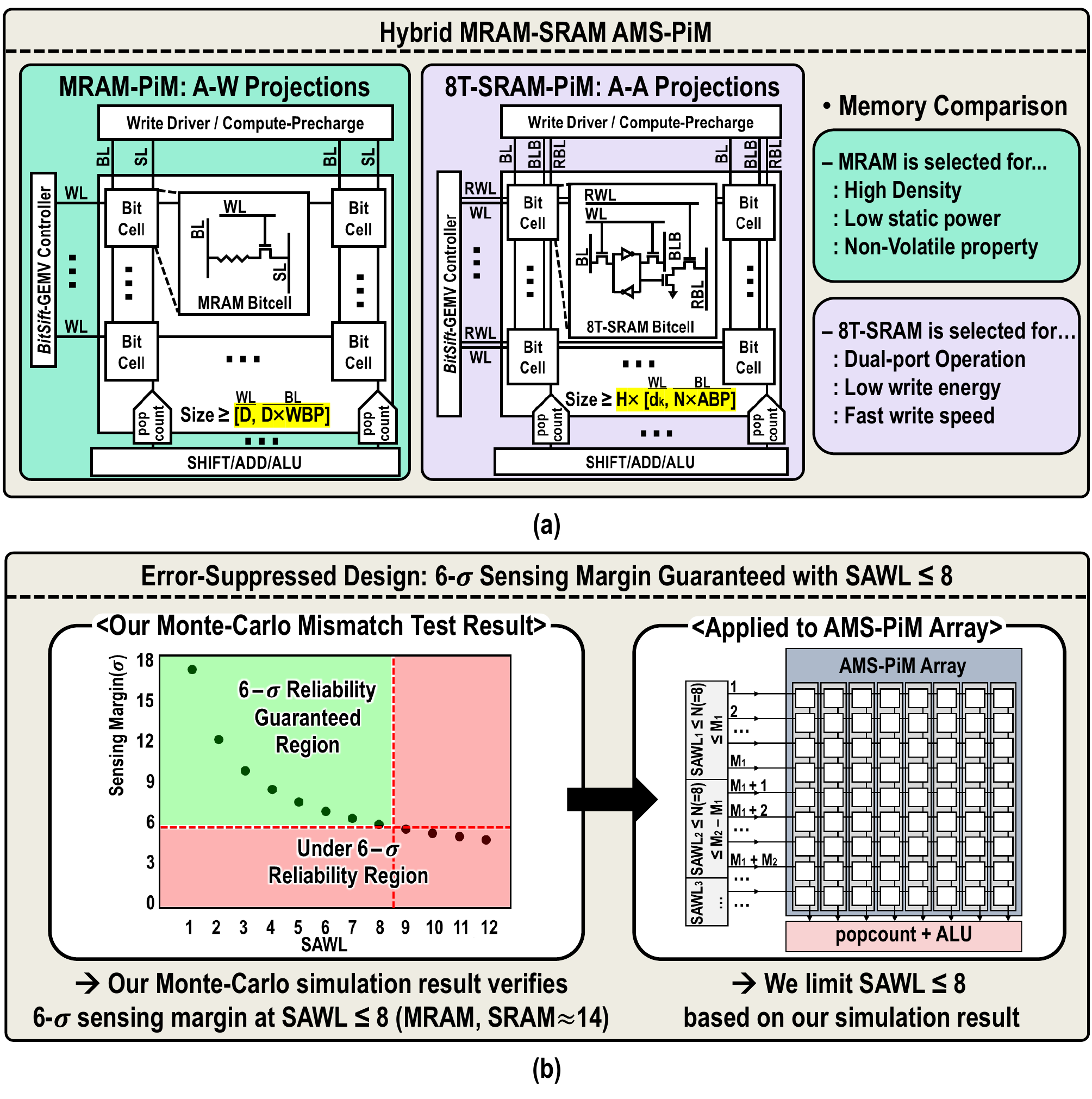}
    \vspace{-6mm}
    \caption{(a) Attribute and size requirements of each memory type, and (b) Monte-Carlo-simulation-guided SAWL selection for lossless computations.}
    \vspace{-4mm}
    \label{fig:MemorySize}
\end{figure}

The basement of our system - MRAM-SRAM hybrid AMS-PiM design - utilizes the intrinsic advantages of each memory for two distinct types of GEMVs: activation-weight (A-W) and activation-activation (A-A) GEMVs, during the DNN's main, and PIM's core; BLAS accelerations.
%Attention layers in transformers require GEMV operations involving both A-W pairs and A-A pairs. 
The MRAM-PiM are well-suited to A-W-pair GEMVs 
%(for Q, K, V, and O generations)
, owing to its non-volatile property which allows for significant weight reuse.  
Furthermore, MRAM maximizes array-level parallelism with a small memory cell size ($\sim$ 1/3 of SRAM). 
On the other hand, A-A-pair GEMVs 
%(for L and A generations)
, which require both operands to be dynamic, are handled using SRAM-PiM devices. 
SRAM offers the advantages of low write energy and high write speed compared to MRAM, making it ideal for A-A GEMVs.

The size and number of arrays are configured to fully encompass all parameters and computations of the self-attention layers, which is a fundamental requirement for achieving the end-to-end acceleration of the target model.
%We also carefully consider our proposed quantization technique, data parallelism, and overall trade-off relations among PPA (power, performance, and area) factors to optimize our proposed architecture. 
The following paragraphs and Fig.~\ref{fig:MemorySize}-(a) outline the minimum array configuration requirements for implementing our FLARE architecture, while our specific design choices are summarized in TABLE~\ref{fig:FATE-PiM_sumamryTable}.

First, the MRAM-CiM arrays for A-W GEMVs should store D$^2$ weight elements for each query-, key-, value-, and the final-output projection layers. 
For each array, the total number of WLs must be larger than the hidden dimension \(D\) so that the input tokens can be full-parallelly fetched, 
and the total number of BLs should be larger than %store the weights of hidden dimensions, but it needs to be 
$D \times WBP$ (weight-bit-precision) to independently store and compute the multi-bit weights at independent BLs. 

Second, for the SRAM-PiM array, %size for A-A GEMVs has different requirements. 
the number of WLs should be guaranteed so that it can 
%can be smaller than that of MRAM arrays, as they are to 
handle $D/H=d_k$ dimensions, while \(H\) independent arrays are needed to handle multi-head projections. 
The columns of the SRAM(s) must store \(N\) weight features, requiring $N \times IBP$ (IBP: input-activation-bit-precision) BLs. 

Meanwhile, to ensure a stable, error-free computation in AMS-PiM devices, 
we limit SAWL$\leq$8 to guarantee stable, 6-$\sigma$ error-free computation, which is verified by Monte-Carlo simulation using our AMS-PiM arrays, where the results are summarized in Fig.~\ref{fig:MemorySize}-(b).
This configuration effectively ensures lossless computation, however, for the latency risk tackled in section~\ref{sec:GEMV_quad_latency}, we propose our fast, accurate, and efficient BitSift-GEMV method in section~\ref{Sec:BitSift}.

\subsection{Dequantization-Free PTQ Technique}\label{sec:fpless_Quantization}

\begin{figure}[h]
    \centering
    \includegraphics[width=\linewidth]{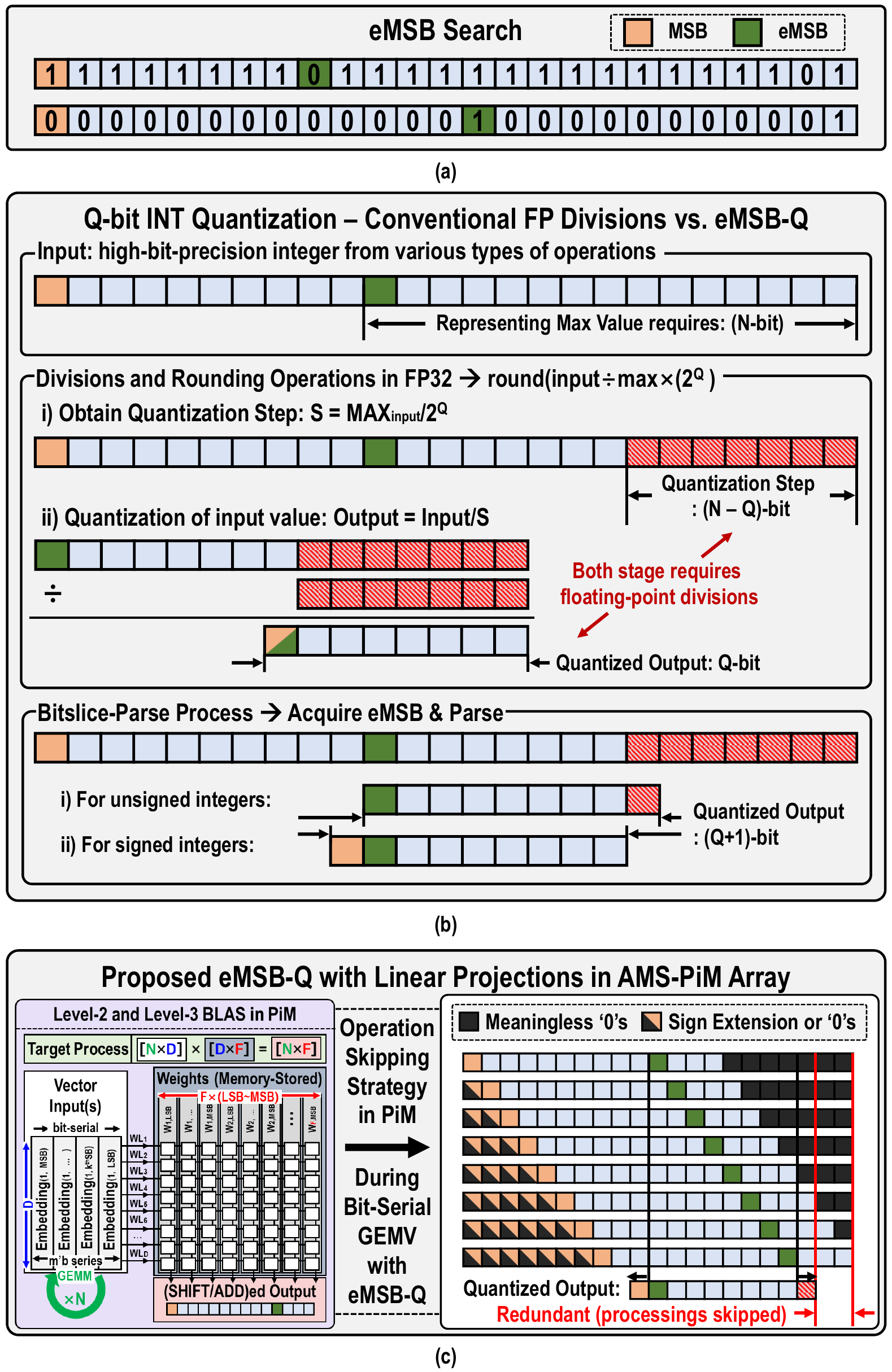} 
    \vspace{-7mm}
    \caption{%How the FP arithmetic operations were replaced in o
    Our FP- and division-less integer quantization method, eMSB-Q. (a) Definition of eMSB for negative and positive (or unsigned) numbers. (b) eMSB-Q achieves lossless quantization at the expense of 1 more bit. (c) When applying our eMSB-Q for the AMS-PiM array, bit-serial inputs also benefit from skipping irrelevant computations.}
    \vspace{-1mm}
    \label{fig:eMSB_Q}
\end{figure} 

%\subsection{Dataflow Demystified}
%QKV gen $\rightarrow$ K and V gen only, and store them at SRAM-PiM
%Stream Q gen $\rightarrow$ catch fly-away to process QK$^T$ for Q from each word, and fuse (quantization $\rightarrow$ Softmax per word vector $\rightarrow$ Attend to V $\rightarrow$ and Final projection) process; so that we don't have to store each of intermediate (Q, QK$^T$, Softmax result, Attended out result) out of Attenete-PiM PE. 
%Streaming on-the-fly Q, QK$^T$, and Softmax outputs reduce not only the memory footprint in inter-device manner but also reduces scratchpad memory required.
%Also, the quadratic bottleneck was induced as we need to fetch n$^2$ vectors and process activation-activation pair GEMVs, however, by storing whole-K-computation result at SRAM-PiM and then processing QK$^T$ by streaming Q results, we can flatten the quadratic bottleneck that induces \(O(n^2)\) tensor footprint to  \(O(2\times n) \approx O(n) \)
%This mitigates the most massive tensor footprint and linearizes the quadratic bottleneck. 

In this section, we propose a dequantization-free PTQ that penetrates all computations within the attention layer, replacing the FP and division arithmetics while retaining accuracy.
%As previously discussed, FPUs or out-of-PiM footprints are known to be crucial for DQ-Q and nonlinear-layer processing. 
%This requirement paralyzes the efficacy of PiMs, with the daunting amount of intermediate tensor traffic. 
%involving FP division arithmetic in 
%The proposed quantization method focuses on (1) removing FPU, (2) complying with the attributes of various computations within the attention layer, ND (3) minimizing the hardware overhead (by eliminating division arithmetics as much as possible). Our focused design seamlessly penetrates through every operation involved within the attention layer.
%Hence, 
As previously discussed, our proposed quantization method enables lossless integer quantization using only shifting and parsing, accommodating important attributes discussed in section~\ref{sec:current_limits_of_quantizationtechnique}. %aims to (1) eliminate FPU, (2) accommodate the requirements of various types of computations within the attention layer, and (3) cooperate with BLAS/nonlinear layer processing blocks. 
%Our design is carefully integrated into every operation involved in the attention layer.

As discussed in section~~\ref{sec:PTQ_nonBLAS_Optimization}, an integer quantization is ``lossless'' when the requirement from equation~\ref{eq:LosslessQuantization} is satisfied, which occurs if the quantization step size is smaller than that derived from floating-point representations.
This implies that the original input can be divided by any \(2^n\), as long as $2^n$$<$$\mathbf{S}$, where \(\mathbf{S}\) represents the quantization step size used in FP-based quantization. 
Therefore, we decided to replace the dequantization and division process, 
%division with our effective-MSB-Detecting Quantization (eMSB-Q) technique, 
as depicted and compared with the conventional method, in Fig.~\ref{fig:eMSB_Q}-(a) and (b).  
Our method, termed effective-MSB-based Quantization (eMSB-Q), ensures numerical integrity while replacing the dequantization and quantization-step acquisition process with eMSB detection, and division operations with arithmetic shifting and parsing. The eMSB-search process simply identifies the location of the actual MSB within a group of bitwise values, eliminating the need for sorting to determine the maximum ``value'' and its ``index.''
%Our method, named effective-MSB-based Quantization (eMSB-Q), retains the numerical integrity by replacing dequantization and quantization-step acquisition by eMSB detection, and division by arithmetic shifting and parsing. 
%Note that our eMSB-search process is simply finding where the actual MSB plays its role along the group of bitwise values; which also alleviates the sorting process for finding out and storing the maximum ``value'' and its ``index''.

To prevent outliers from being clipped during eMSB-Q, we apply per-token quantization for \(\mathbf{Q}\), \(\mathbf{QK}^T\), and the final output. 
This approach leverages that each token vector in the \(\mathbf{Q}\)-matrix is independently processed throughout the self-attention mechanism. 
%While applying eMSB-Q, to prevent outliers from being clipped, we apply the quantization process per token (of Q, QK$^T$, and final output generation). 
%The per-token quantization stems from the knowledge that the vector representing each token in the Q-matrix is independently processed along the self-attention process. 
However, KV generations - which must preserve global context - involve storing all vectors of \(N\) tokens. 
It differentiates the parsing strategy: we parse bits starting from MSBs rather than eMSBs, using a longer bit to reserve a wider dynamic range for the global context.

Note that, the inherent quality of linear operations is kept with eMSB-Q.
However, to save the distinct attributes within the nonlinear layer using our per-token quantization process, we transfer the eMSB information to the VDR-Softmax block, enabling it to incorporate exponent data during processing. This step is essential for the precise functioning of the Softmax layer, which relies on accurate exponent handling, unlike linear layers.
In the following section, we further describe integer-processed Softmax for PTQ.

\begin{algorithm}[t]
\small
\caption{VDR-Softmax}
\begin{algorithmic}\label{alg:iSoftMax}
\STATE \textbf{Input:} $x$ [Q$_I$-1:0][(N-1):0], \textit{n$_e$}
\STATE N = number of the input elements
\STATE Q$_I$ = input bit precision, Q$_O$ output bit precision
\STATE \textit{n$_e$} = $\sum$(MSB-eMSB) over input$\rightarrow$Q$\rightarrow$QK$^T$
\STATE \textbf{Parameters:}
\STATE $~ ~ a, b, c, S, l \gets a(n_e), b(n_e), c(n_e), S(n_e), l(n_e)$
\STATE $~ ~\Rightarrow \text{once for } Q_I, \text{incorporates exponent information}.$ 
\STATE $~ ~\Rightarrow \text{LUT-processed}, 100\text{ bytes}$
\STATE \textbf{Output:} $x_{Softmax}$ [Q$_O$-1:0][(N-1)):0] 
\STATE \text{ }
%%%
\STATE \textbf{function} VDR\_iPOLY(r, \textit{n$_e$})
\STATE $~ ~ ~ ~x_{POLY} \gets r\times (r + b) + c$
\STATE $~ ~ ~ ~S_{POLY} \gets a\times S$
\STATE $~ ~ ~ ~\textbf{return}~x_{POLY}, S_{POLY}$
%\STATE \textbf{end function}
\STATE \text{ }
%%%
\STATE \textbf{function} VDR\_iEXP(x$_{sub}$, \textit{n$_e$})
\STATE $~ ~ ~ ~Q \gets \text{clip}(int(X_{sub} / l), 2\times Q_I)$
\STATE $~ ~ ~ ~r \gets x_{sub} - Q - l$
\STATE $~ ~ ~ ~r_{POLY}, S_{POLY} \gets \text{iPOLY}(r, S)$
\STATE $~ ~ ~ ~r_{EXP} = r_{POLY} << Q$
\STATE $~ ~ ~ ~S_{EXP} = S_{POLY} >> Q$
\STATE $~ ~ ~ ~\textbf{return}~r_{EXP}, S_{EXP}$
%\STATE \textbf{end function}
\STATE \text{ }
%%%
\STATE \textbf{function} VDR\_Norm(x, S)
\STATE $~ ~ ~ ~x_{sub} = x - \text{max}(x)$
\STATE $~ ~ ~ ~r_{EXP}, S_{EXP} = iEXP(x_{sub}, S)$
%\STATE $~ ~ ~ ~Q_{U} = (c << 2\times Q_I) >> Q_O$
%\STATE $~ ~ ~ ~x_{Softmax} = r_{EXP}$
\STATE $~ ~ ~ ~x_{Softmax} = \text{eMSB-Q}(r_{EXP})$
\STATE $~ ~ ~ ~\textbf{return}~x_{Softmax}, S_{EXP}$
%\STATE \textbf{end function}
\end{algorithmic}
\end{algorithm}

\subsection{Proposed VDR-Softmax Fused with eMSB-Q}\label{sec:VDRsoftmax}

The simplified yet accurate implementation of the Softmax layer in PTQ requires a precise understanding of its characteristics and a more sophisticated integer-processing technique. 
The implementation involves several key steps, outlined in the Algorithm~\ref{alg:iSoftMax}. 
Our approach utilizes per-token eMSB information to ensure accurate yet FP-less processing and uses only shifting and parsing of integers to normalize and quantize the score into integer format. 

While accommodating exponent information in processing the nonlinear layer is critical, as visualized in Fig.~\ref{fig:CurrentLimit_of_Quantization}-(c), directly multiplying/dividing the integer $x$ makes the number exceed the representable range of the integers with designated bit-precision, leading to loss of information. 
The key to handling Softmax processing with integer arithmetic lies in adapting the exponent indirectly while approximating the $e^{x/n}$ function, where $x$ is an integer-format input value and $n$ is a representation of the exponent. 
To accomplish that, we adjust the base ($e$) of the exponential function from $e$ to $^n\sqrt{e}$, rather than adjusting the integer-represented $x$ in order to avoid the loss of information. 
Using well-established, $e^x$-approximating functions from previous works~\cite{kim2021bert,li2023vit}. 
This adjustment is accomplished by modifying the internal parameters of the proposed method, without requiring division arithmetic.

Additionally, the base adjustment along the exponentiation process - from $e$ to $^n\sqrt{e}$ - is performed per token, using eMSB data from the eMSB-Q stage.  
Since Softmax is computed per token, aligned with our quantization approach (in section~\ref{sec:fpless_Quantization}) and dataflow control mechanism (in section~\ref{sec:ReducedTensorFootprints}),
we do not need to store eMSB positions for all Q vectors, enhancing our architecture's scalability. 

After the exponentiations, we apply eMSB-Q to a group of exponentiated outputs for each token, maximizing the utilization of the dynamic range provided by a Q$_{O}$-bit (Softmax output precision) integer. This approach avoids division while preserving the original proportions of the values, ensuring that the maximum value is mapped to the highest possible representation, with sufficient range allocated for smaller values.%achieves an efficient and sufficiently accurate quantization of the Softmax output.
%We validated that our proposed Softmax method works better with a wider dynamic range of input with extensive simulations, as shown in \textcolor{red}{Table XXX}.

%%%%%% Detailed Computation Steps %%%%%%
%1. **Subtraction**: Each element \(x_i\) is adjusted by subtracting the maximum value in \(x\) to maintain numerical stability.
%2. **Segmentation**: The adjusted vector \(x\_sub\) is segmented into smaller chunks for parallel processing.
%3. **Division**: Each segment is divided by a scaling factor to normalize the values.
%4. **Quotient and Remainder Calculation**: These are used to facilitate the polynomial approximation of the exponential function.
%5. **Polynomial Approximation**: An integer-based polynomial approximation is applied to compute the exponentials efficiently.
%6. **Exponent Calculation**: The computed exponentials are scaled and adjusted to fit within the desired output range.
%7. **Softmax Calculation**: The final Softmax values are obtained by normalizing the exponentials, ensuring that they sum to one.

Consequently, our VDR-Softmax achieves both efficiency and accuracy, making it well-suited for end-to-end, integer-based acceleration within our PTQ-based attention accelerator architecture. This approach not only retains the critical properties of the exponentiation and divisions in Softmax, but also alleviates the computational overhead, leading to significantly enhanced efficiency with secured performance.

\subsection{BitSift-GEMV Processor}\label{Sec:BitSift}

Along the end-to-end, on-device fusion of the self-attention layer, our BitSift-GEMV technique for AMS-PiM significantly boosts sparse GEMV, which composes most of the computations in the attention layer. BitSift-GEMV is motivated by two key factors: \textbf{(1)} the highlighted bitwise sparsity of activation data illustrated in Fig.~\ref{fig:comparison_various_sparsity} and \textbf{(2)} the constant-maximum number of ``1''s (=8) WL activations, which allows the highest-utilization-rate operation and strengthened robustness of ADC designs.

\begin{figure}[t]
    \centering
    \includegraphics[width=\linewidth]{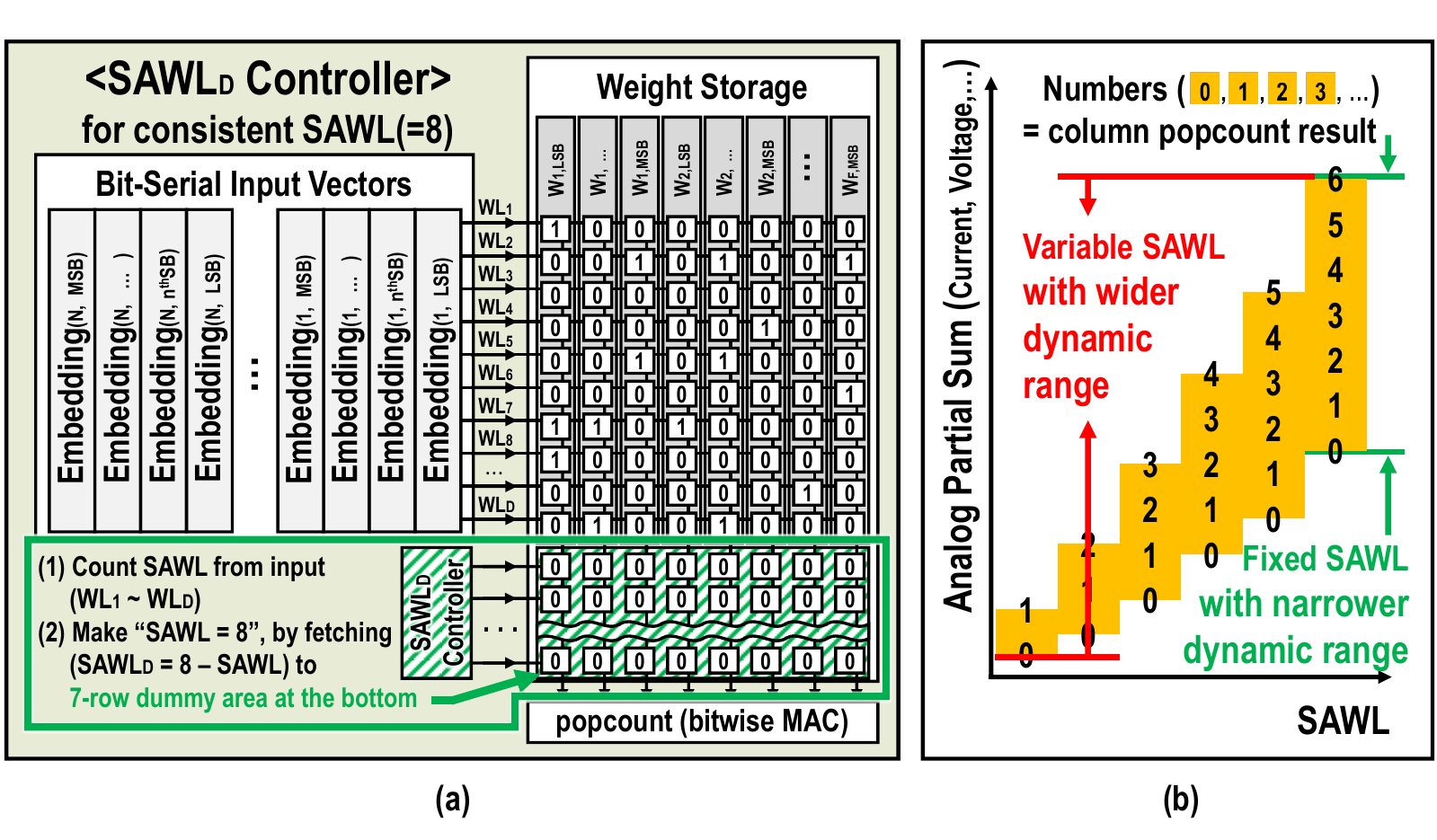}
    \vspace{-5mm}
    \caption{(a) SAWL$_D$ controller enabling a fixed number of SAWL(=8) when the input data is insufficient of ``1''s. (b) Relation between SAWL and the dynamic range variation of analog-signal-domain popcount. Limiting flexibility with SAWL relaxes the ADC-design challenges for AMS-PiM.}
    \vspace{-4mm}
    \label{fig:SAWL_D_Control}
\end{figure}

Existing AMS-PiM designs require flexibility in the number of processible SAWL as input vectors have a random number of ``1''s along the bit-slice. 
One thing to note here is that 
a flexible SAWL can lead to worse computation reliability with higher overhead for the ADC design (Fig.~\ref{fig:SAWL_D_Control}-(b)). 
For SAWL-flexible configurations, the ADCs should handle a wider, variable dynamic range, higher resolution, and varying threshold during the analog-to-digital conversions. 
This also compromises the utilization rate when the actual SAWL lowers, wasting power/area implemented for higher SAWL. 
Therefore, our design sets ``SAWL=MAX'' for all time without flexibility, by incorporating dummy-``1'' inputs to maintain a fixed, maximum number of ``1''s.

To maintain a constant number of SAWL at 8 (= maximum SAWL that allows 6-$\sigma$ reliability), %rather than allowing flexibility from 0 to 8, 
the SAWL$_D$ controller in Fig.~\ref{fig:SAWL_D_Control}-(a) supplements the SAWL with additional ``1''s, when the input vector has fewer  ``1''s than 8. 
Specifically, the SAWL$_D$ controller activates SAWL$_D$ = (8$-$SAWL) WLs at the bottom of the array.
This is achieved by incorporating a dummy array and WL interfaces consisting of seven rows at the bottom. 
The memory cells within these dummy rows are all set to ``0'' (off-cells), emulating computations with dummy inputs. 
The array configuration serves two purposes: (1) isolating dummy operations - it prevents the dummy array from affecting the actual computation output, and (2) enhancing variation resiliency - by consistently returning ``0''s, the dummy array area improves the system's robustness against variations. 
%The memory cells in dummy rows are all filled with ``0''s (off-cells) to detach their effect of operation from the computation output and to ensure better variation resiliency, by making the dummy array area always return ``0''s. 
I.e., While processing only ``1''s along the input vector, the bitwise ``0''s along the vector(s) are skipped from the actual computation. 

The fixed-SAWL processing incorporating SAWL$_D$ controller employs a two-step approach.
First, the input vector is scanned to identify the longest slice containing eight ``1''s. The slice is then selected for processing. 
Second, for the tail end of a vector with fewer ``1''s (or, for a very sparse input vector), the SAWL$_D$ processor supplements the existing ``1''s with additional ones from the dummy array. This ensures a consistent maximum SAWL for all computations, regardless of the input vector's density.
By maintaining a constant SAWL of 8, this method fixes the input dynamic range for ADC and enables lower area/energy/error overhead, even with varying input sparsity. 
%This contributes to the system's overall efficiency and accuracy.

%With the proposed processing mechanism, we scan through the input vector to collect 8-``1''s along the vector to fetch the longest slice of the input vector with SAWL=``8'', before fetching the input to the WLs. For a very sparse vector or the last section of the input vector that has less ``1''s, the SAWL$_D$ processor suffices ``1'' to ensure consistent AMS computations.
%Since the deactivated WL cells do not affect the computation result, we can fetch whatever number of ``0''s along the input to the WLs. 

\begin{figure}[]
    \centering
    \includegraphics[width=\linewidth]{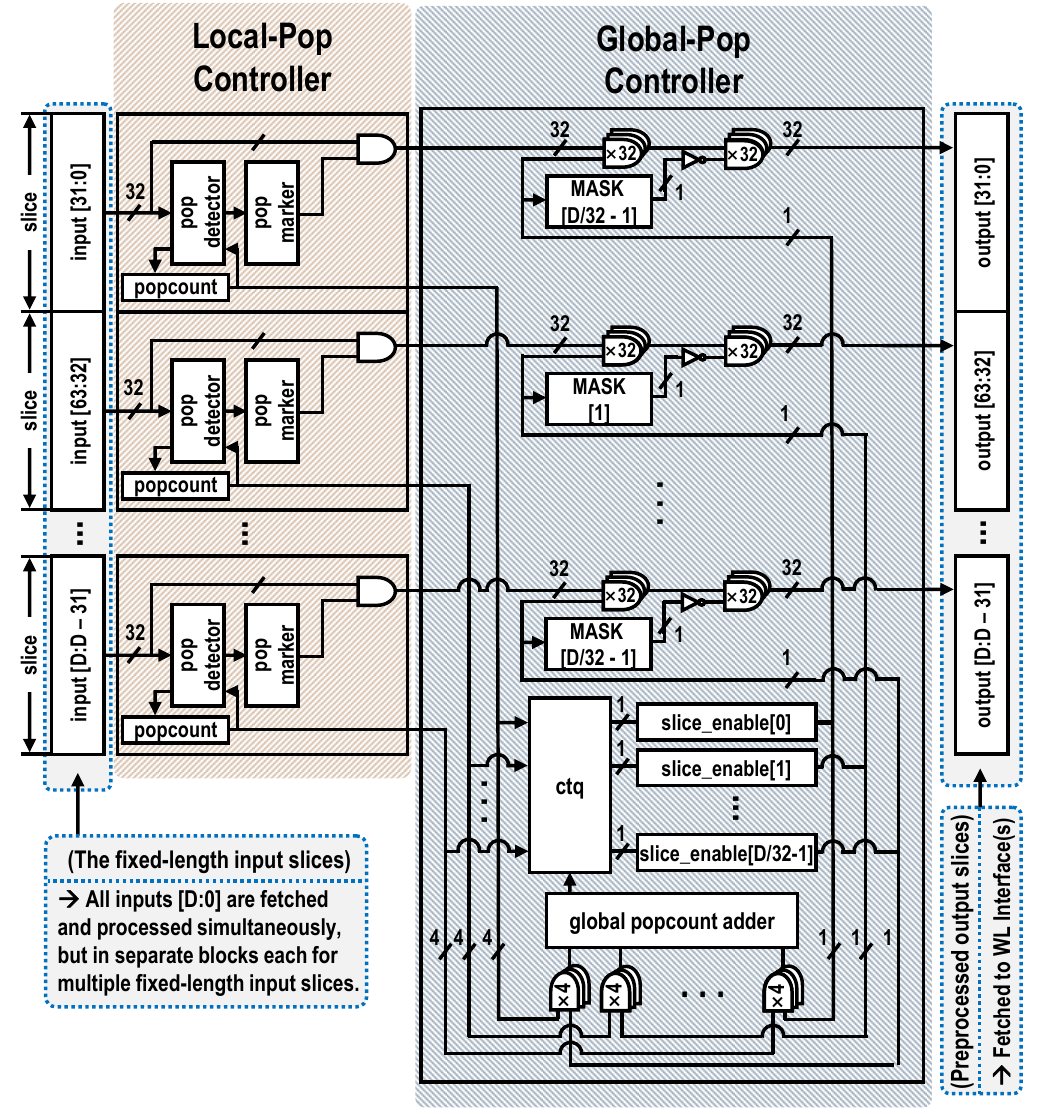}
    \vspace{-4mm}
    \caption{Our BitSift-GEMV controller parses, selects, and fetches the longest portion(s) of the input - that contains designated SAWL - to WL interface(s). %While hiding relatively faster logic operations at AMS-GEMV operations, we can achieve faster and more efficient GEMV compared to both digital counterparts and AMS-PiM baselines.
    }
    \vspace{-4mm}
    \label{fig:BitSiftController}
\end{figure}

To support the SAWL$_D$ controller by identifying the longest segment(s) of the input with the largest number of SAWL that remain constrainted of being $\leq$8, we introduce our BitSift-GEMV controller, depicted in Fig.~\ref{fig:BitSiftController}. 
%A key contribution of this controller is its novel mechanism for identifying and collecting the longest segment of the input vector containing eight "1"s. 
This strategy introduces a novel approach for highly efficient sparse GEMV processing, achieving low latency and minimal area/energy overhead. Instead of a hypothetical brute-force method that might scan the entire input vector bit by bit - resulting in significant latency penalties as vector length and sparsity increase - the BitSift-GEMV controller leverages a hierarchical design with high parallelism. This includes 
(1) a local-pop controller (LPC) and (2) a global-pop controller (GPC), where ``pop-count'' refers to the number of ``1''s in the input vector.

\textbf{(1)} The LPC efficiently counts the number of ``1''s within short, fixed-length slices of the input vector using dedicated ``pop detector'' and ``popcount'' blocks. This localized processing ensures low-latency computation of the popcount within each slice.
\textbf{(2)} The GPC aggregates the popcounts from LPCs to rapidly identify the highest possible number of ``1''s, constrained to a maximum of 8, for further processing. Once aggregation is complete, the GPC activates the ``MASK'' to fetch the corresponding vector to the output and WLs.
If the total number of ``1''s identified by the GPC is fewer than 8, the SAWL$_D$ controller (Fig.~\ref{fig:SAWL_D_Control}-(a)) supplements the input by adding ``1''s from the dummy array before forwarding it to the WLs of PiM array. 
This hierarchical and parallel architecture enables our AMS-PiM arrays to sustain high performance, even as input vector lengths and sparsity increase.
%The LPCs operate independently and concurrently on their assigned input slices, enabling significant parallelism and reducing overall latency. 
The design is specifically optimized for extremely sparse inputs, especially whose popcount within each LPC's slice remains $\leq 8$. 
%The design is optimized for extremely sparse inputs, specifically for input vectors with popcount within each LPC's slice remain $\leq$8.
This emphasis on sparsity aligns with the activation values frequently observed in neural networks, as illustrated in Fig.~\ref{fig:comparison_various_sparsity}-(e), (f).

However, recognizing that real-world data can exhibit varying sparsity patterns, the BitSift-Controller also incorporates dedicated circuitry to handle denser input segments. 
That is, 
our BitSift-GEMV controller dynamically adjusts the processing flow when an LPC encounters eight or more ``1''s within its slice, or when the GPC detects a global popcount exceeding 8. This dynamic adjustment mechanism is as follows. 
When the ``pop detector'' block in LPC detects 8$\times$``1''s before reaching the end of the sliced input, the inputs are marked from the beginning to the point where the 8-th ``1'' is found, using the ``pop marker'' block. 
The marked section is then fetched first, consuming a column-sum cycle in the AMS-PiM array. 
After that, the marker marks from the next section of the inputs to be processed - excluding the section fetched so far - and continues until the process reaches the end of the LPC's processible input section.
After the LPC-level sparsity requirement (popcount$<$8) is all met, 
in the GPC, when the global popcount adder returns $>$8, the ctq (compute-token queue) circuit block controls the ``MASK''s, gating the LPCs from fetching the input to the PiM array so that the LPCs of having summed popcount under 8 can be only fetched. 
Thus along the input vector of length [D], it takes processing cycles up to roundup(${D\times\text{(1-bitwise\_sparsity)}}\over8$). 
Therefore, our BitSift-GEMV technique reduces the processing latency by (1 $-$ bitwse$\_$sparsity), fully utilizing the finest grain of sparsity.

\subsection{Reduced Tensor Traffic}\label{sec:ReducedTensorFootprints}
\begin{figure}[h]
    \centering
    \vspace{-3mm}
    \includegraphics[width=0.97\linewidth]{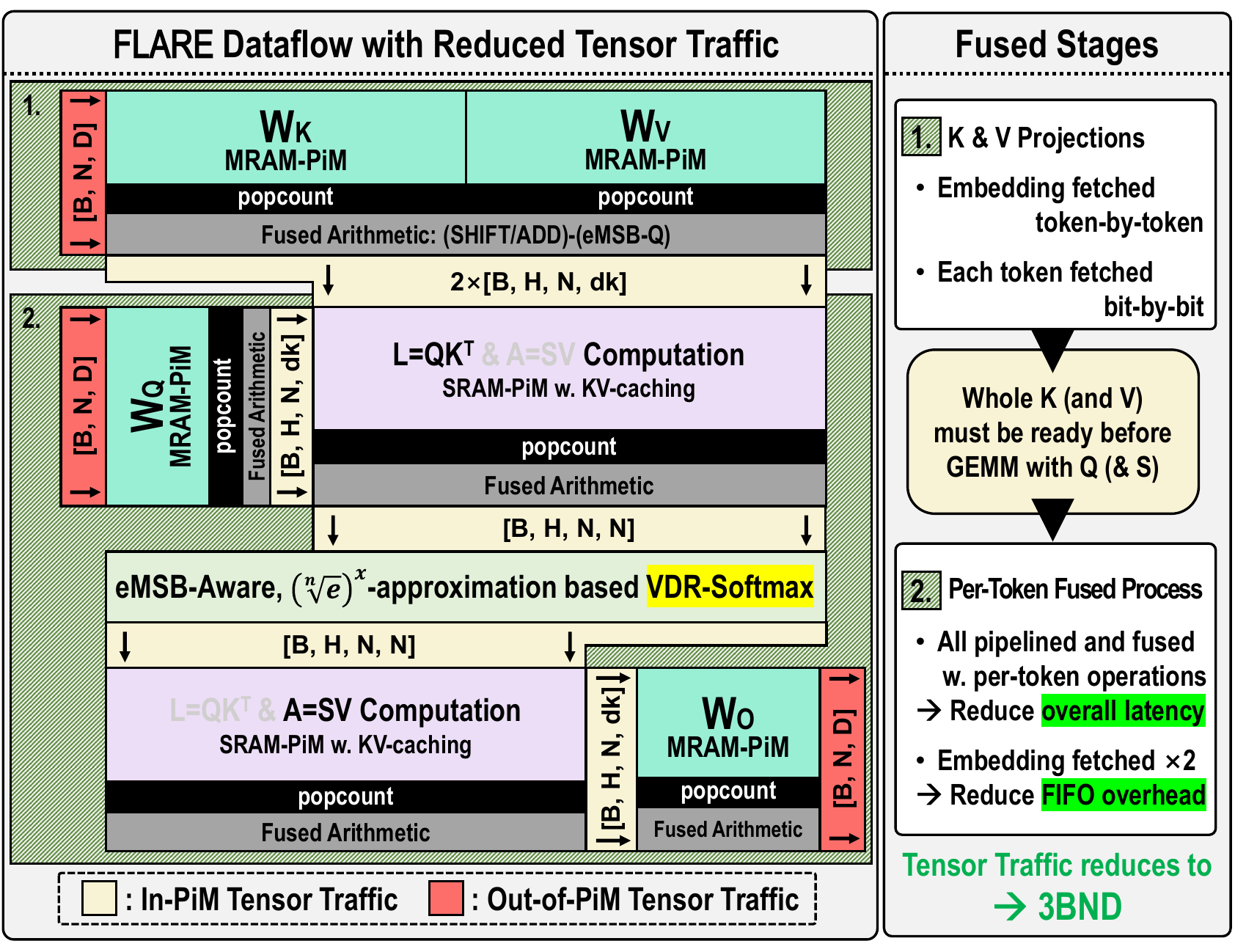}
    \vspace{-2mm}
    \caption{Visualized Dataflow and tensor traffic of FLARE: QKV generation, on-the-fly processing, and fused operations.}
    \vspace{-4.5mm}
    \label{fig:Atten_PiM_Dataflow}
\end{figure}
Conclusively, our FLARE architecture minimizes tensor traffic in attention mechanisms by fusing the end-to-end attention layer with combined novel techniques. 
The dataflow of our FLARE architecture (depicted in Fig.~\ref{fig:Atten_PiM_Dataflow}) can be understood as follows: 
(1) K \& V Projections and
(2) end-to-end fused, per-token (\(Q\)-\(L\)-\(A\)-\(O\)) process. 

(1) K \& V projections: This stage involves A-W GEMM (a series of GEMV) operations on input tokens with weight matrice W$_K$ \& W$_V$, requiring input streaming to generate the complete KV matrices.

(2) Per-token fused process: After generating KV matrices, the end-to-end-fused attention process is handled, requiring a second input stream-in. Rather than storing an input in an internal buffer, we stream data twice for improved area efficiency and scalability.
%The total latency is approximated as the latency required for (1) + (2), as from the second stage, the final output is processed with a fixed small amount of delay in a per-token manner.

%The proposed Atten-PiM architecture reduces the tensor footprint by efficiently fusing operations and minimizing intermediate storage. 
Hence, the revised tensor traffic can be represented as:
\[
\text{Revised Tensor Traffic} = 2 \times B \times N \times D \text{ (in)} + B \times N \times D \text{ (out)}~,
\]
%The proposed architecture 
achieving substantial tensor traffic reduction by leveraging our proposed techniques. %MRAM-SRAM hybrid PiM arrays, efficient data caching, fused arithmetic operations, and reduced FIFO overhead. These strategies collectively enable a more compact and efficient attention processing, crucial for the scalability and performance of transformer models.

%%%%%%%%%%%%%%%%%%%%%%%%%%%%%%%%%%%%%%%%%%%%%%%%%%%%%%%%%%%%%%%%%%%%%%%%%%%%%%%%%%%%%%%%%%%%%%%%%%%%%%%%%%%%%%%%%%%%%%%%%%%%%%%%%%%%%%%%%%%%%%%%%%%%%%%
%%%%%%%%%%%%%%%%%%%%%%%%%%%%%%%%%%%%%%%%%%%%%%%%%%%%%%%     N E W    S E C T I O N     %%%%%%%%%%%%%%%%%%%%%%%%%%%%%%%%%%%%%%%%%%%%%%%%%%%%%%%%%%%%%%%%
%%%%%%%%%%%%%%%%%%%%%%%%%%%%%%%%%%%%%%%%%%%%%%%%%%%%%%%%%%%%%%%%%%%%%%%%%%%%%%%%%%%%%%%%%%%%%%%%%%%%%%%%%%%%%%%%%%%%%%%%%%%%%%%%%%%%%%%%%%%%%%%%%%%%%%%

\section{FLARE Evaluations}  
\begin{table}[h]
    \centering
    \vspace{-2mm}
    \caption{Summary of our FLARE design in 28nm process}
    \vspace{-4mm}
    \includegraphics[width=\linewidth]{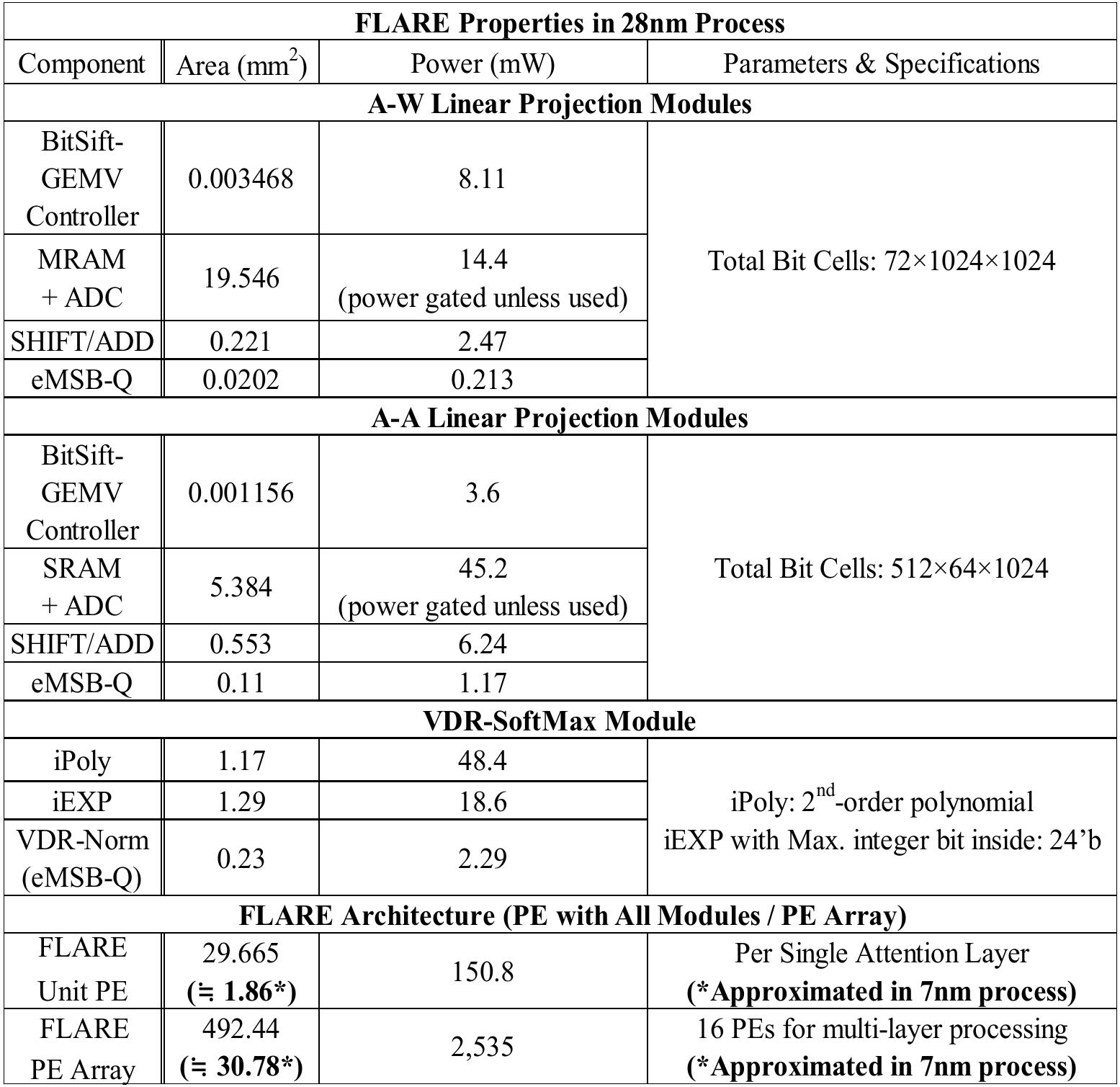}
    \vspace{-4mm}
    \label{fig:FATE-PiM_sumamryTable}
\end{table}
\noindent Our proposed FLARE architecture and its design components are all embedded into a compact ASIC with superior energy and latency performance per token. We validate our design through meticulous circuit simulations, including Monte Carlo simulations (result summarized in Fig.~\ref{fig:MemorySize}-(b) and
post-layout circuit simulations for physical-level feasibility, using our custom-designed ASIC on 28nm FD-SOI technology, and our design is summarized in TABLE~\ref{fig:FATE-PiM_sumamryTable}.

%%%%%%%%%%%%%%%%%%%%%%%%%%%%%%%%%%%%%%%%%%%%%%%%%%%%%%%%%%%%%%%%%%%%%%%%%%%%%%%%%%%%%%%%%%%%%%%%%%%%%%%%%%%%%%%%%%%%%%%%%%%%%%%%%%%%%%%%%%%%%%%%%%%%%%

We tested our FLARE using 8-bit %weight and (8+1)-bit input activation (+1 for internal eMSB-Q, baselines with other hardware were quantized to 8-bit) 
integer-quantized models across different NLP (with GLUE benchmarks~\cite{wang2018glue}) and vision (ImageNet~\cite{deng2009imagenet} Classification) tasks, as well as with various transformer models~\cite{touvron2021training,dosovitskiy2020image,liu2019roberta,devlin2018bert}. We compared the results with state-of-the-art GPUs (Nvidia RTX 3090, RTX 4090, T4, and A100), as well as a PiM baseline where we excluded our proposed techniques. 

Our experimental results validate the FLARE architecture’s diverse efficacy. The design exploration highlights the significance of customizing inner-array configurations and parallelism levels to optimize performance. The eMSB-Q and VDR-Softmax techniques maintain high inference accuracy while eliminating FP operations, thereby enhancing computational efficiency. Additionally, the BitSift-GEMV technique leverages bitwise sparsity to significantly accelerate GEMV operations. Collectively, these innovations establish our architecture as a robust and scalable solution for transformer acceleration, delivering substantial reductions in both latency and energy consumption.

\subsection{Hardware Customization Exploration}\label{sec:HW_design_Exploration}
\begin{figure}[h]
    \centering
    \vspace{-4mm}
    \includegraphics[width=\linewidth]{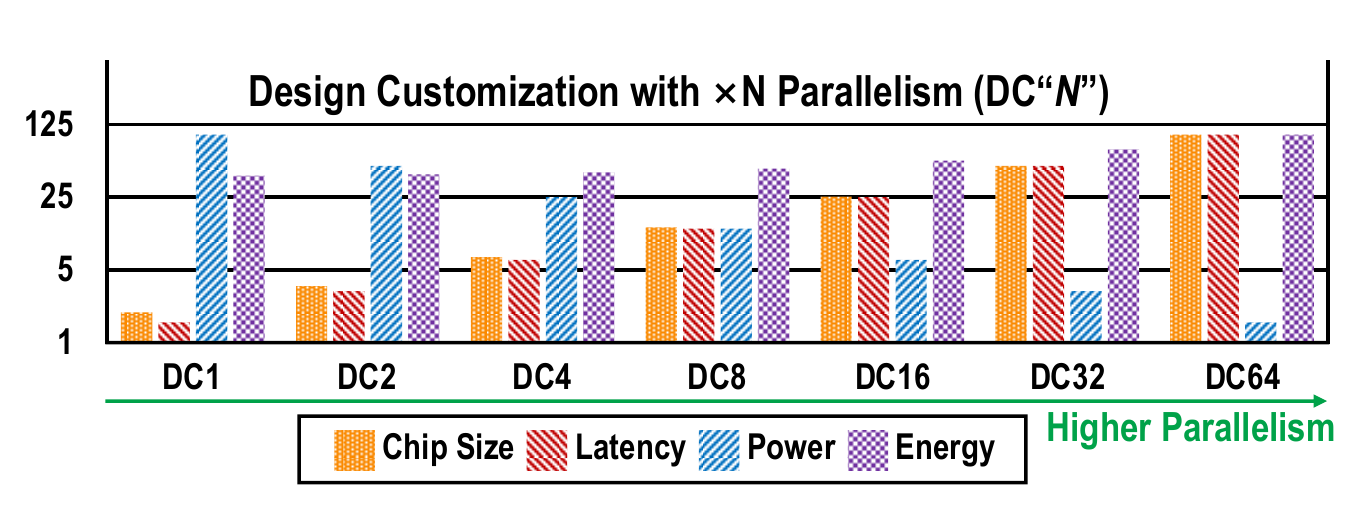}
    \vspace{-6mm}
    \caption{Design study on array-level device parallelism. Larger values imply better characteristics.}
    \vspace{-2mm}
    \label{fig:Experiment_DesignExplorations}
\end{figure}
We conducted a design exploration study to understand how different array-parallelism configurations affect the overall performance of our proposed architecture. Fig.~\ref{fig:Experiment_DesignExplorations} shows the normalized metrics of various design choices, including chip size, latency, power, and energy consumption. For the summary in Table~\ref{fig:FATE-PiM_sumamryTable}, our choice for the parallelism is 8$\times$ array (i.e., DC8), to show the design result with a medium benchmark. 

%%%%%%%%%%%%%%%%%%%%%%%%%%%%%%%%%%%%%%%%%%%%%%%%%%%%%%%%%%%%%%%%%%%%%%%%%%%%%%%%%%%%%%%%%%%%%%%%%%%%%%%%%%%%%%%%%%%%%%%%%%%%%%%%%%%%%%%%%%%%%%%%%%%%%%

\subsection{Impact of eMSB-Q and VDR-Softmax}
\begin{figure}[h!]
    \centering
    \vspace{-3mm}
    \includegraphics[width=\linewidth]{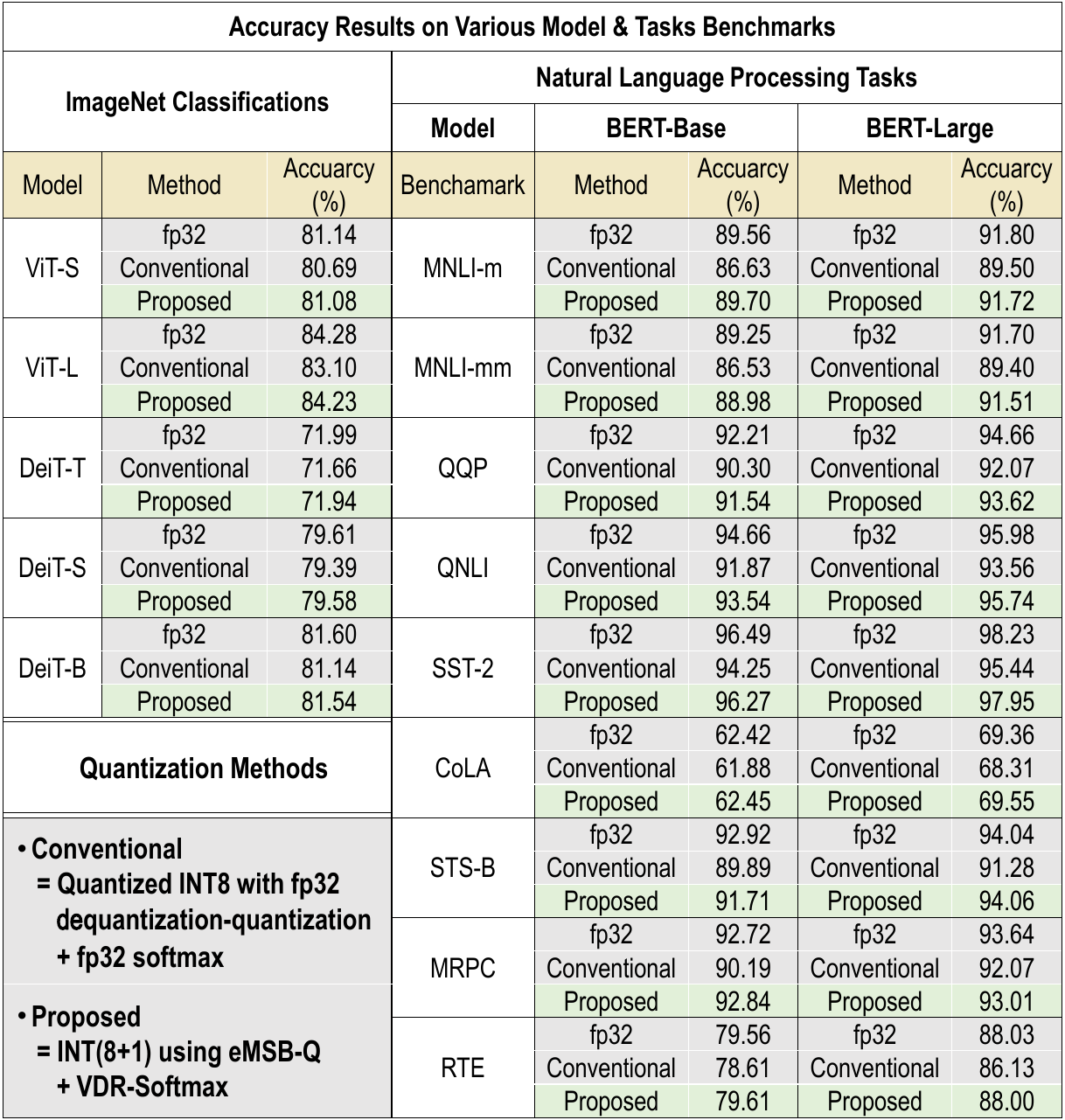}
    \vspace{-6mm}
    \caption{Comparison of inference accuracy results: various NLP tasks and ImageNet classification tasks with different transformer models.}
    %{★!!FOR OUR INSIGHT!!★}% When input values are `actually' quantized in integer format (0000, 0001, ...) than just having discrete, smaller number of levels in FP32 (0.0001, 0.0002, ...), the accuracy fall was larger than the results reported from 8-bit quantization works... GPUs may use int8 with fake-quantization, however, PiM CANNOT!!! (when we look into actual code of various quantization implementations, they surprisingly are!!) eMSB-Q + VDR-Softmax resolves this issue as was proposed.
    \vspace{-2mm}
    \label{fig:comparison_NLPAccuracy}
\end{figure}
Fig.~\ref{fig:comparison_NLPAccuracy} presents the inference accuracy benchmarks across various tasks and models. The comparison includes the FP32 baseline, a traditionally-integer-quantized model utilizing the FP32-based DQ-Q process, and our proposed method employing eMSB-Q quantization combined with the VDR-Softmax technique. The results demonstrate that our approach sustains high accuracy while eliminating FP operations, thereby achieving substantial improvements in computational efficiency.

%%%%%%%%%%%%%%%%%%%%%%%%%%%%%%%%%%%%%%%%%%%%%%%%%%%%%%%%%%%%%%%%%%%%%%%%%%%%%%%%%%%%%%%%%%%%%%%%%%%%%%%%%%%%%%%%%%%%%%%%%%%%%%%%%%%%%%%%%%%%%%%%%%%%%%
\subsection{Impact of BitSift-GEMV}

\begin{figure}[h!]
    \centering
    \vspace{-5mm}
    \includegraphics[width=0.8\linewidth]{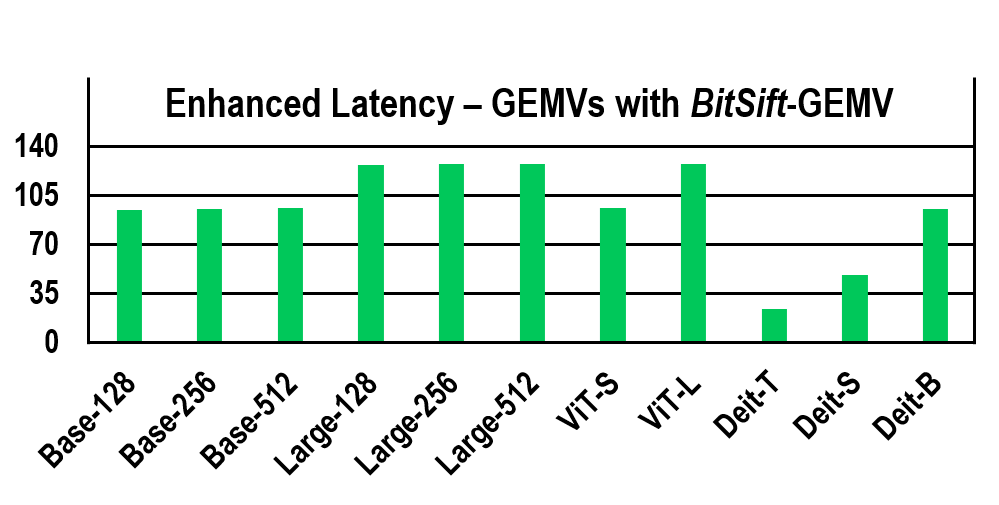}
    \vspace{-3mm}
    \caption{Average amount of how much GEMV operations were boosted. The boosting factors were almost directly the same as our anticipated values, revealing our BitSift-GEMV's efficacy.}
    \vspace{-2mm}
    \label{fig:ImpactofSparsity}
\end{figure}
Fig.~\ref{fig:ImpactofSparsity} presents the impact of our proposed BitSift-GEMV technique on the average number of boosting factors in GEMV operation cycles, compared to PiM baseline with fixed-length processed GEMV. 
Note that the boosting is marked without the effect of array parallelism, and the measurement result is solely boosted by the BitSift-GEMV only. 
The boosting factor observed correlates closely with our anticipated values, demonstrating the effectiveness of exploiting bitwise sparsity to enhance GEMV processing speed.

%%%%%%%%%%%%%%%%%%%%%%%%%%%%%%%%%%%%%%%%%%%%%%%%%%%%%%%%%%%%%%%%%%%%%%%%%%%%%%%%%%%%%%%%%%%%%%%%%%%%%%%%%%%%%%%%%%%%%%%%%%%%%%%%%%%%%%%%%%%%%%%%%%%%%%
\subsection{Impact on Latency Performance}
\begin{figure}[h!]
    \centering
    \includegraphics[width=\linewidth]{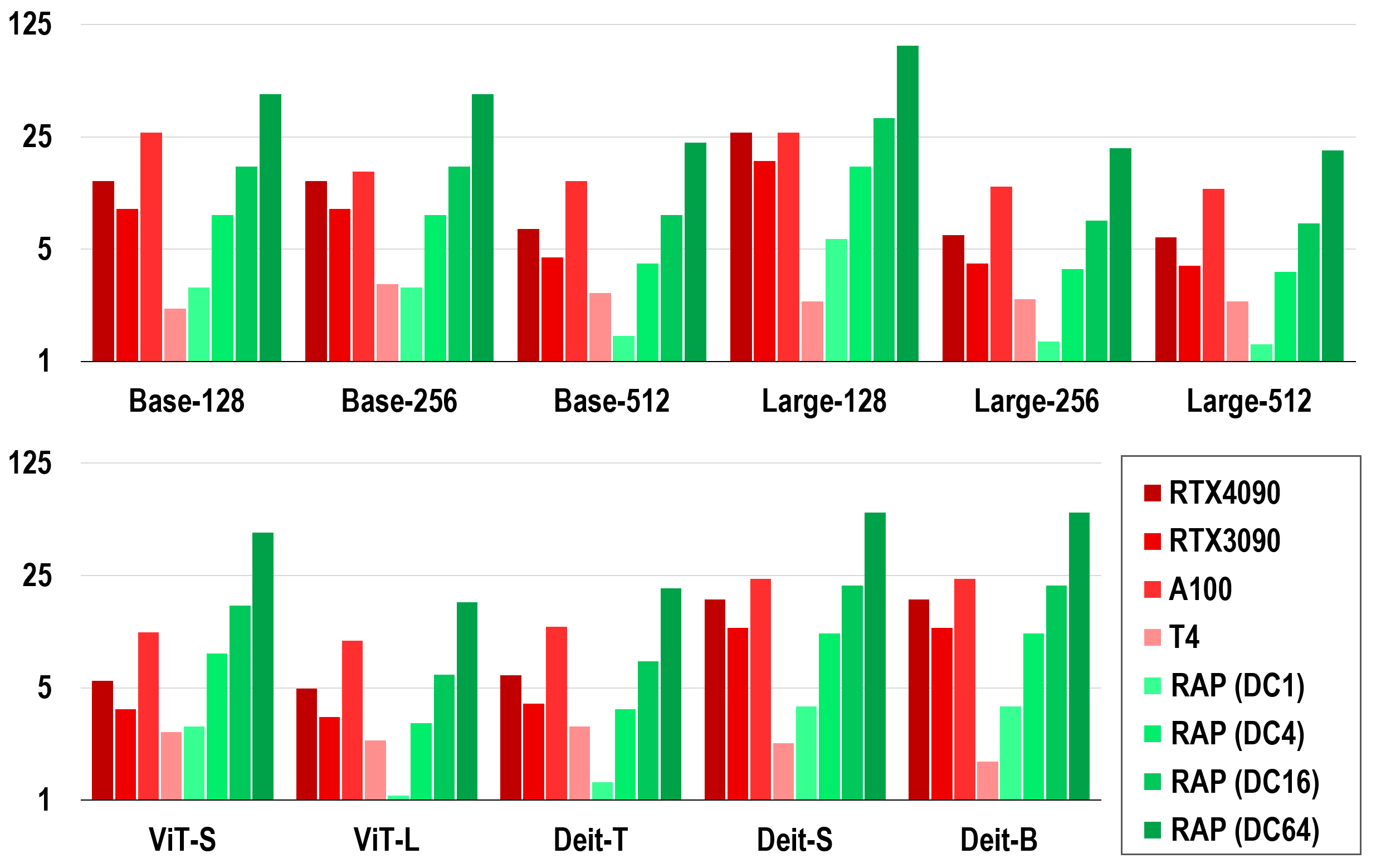}
    \vspace{-5mm}
    \caption{Comparison of normalized token/sec performance. The CiM Baseline was omitted due to its poor performance from significant latency with out-of-PiM tensor traffic and segmented GEMV.}
    \vspace{-2mm}
    \label{fig:comparison_Latency}
\end{figure}
We measured and compared the token/sec performance with various tasks, models, and hardware. 
Fig.~\ref{fig:comparison_Latency} shows that our design achieves competitive latency performance, outperforming not only the PiM baseline but also famous SOTA GPUs. 
This improvement is primarily attributed to the integration of the end-to-end fusion technique and the BitSift-GEMV approach.%%%%%%%%%%%%%%%%%%%%%%%%%%%%%%%%%%%%%%%%%%%%%%%%%%%%%%%%%%%%%%%%%%%%%%%%%%%%%%%%%%%%%%%%%%%%%%%%%%%%%%%%%%%%%%%%%%%%%%%%%%%%%%%%%%%%%%%%%%%%%%%%%%%%%%

\subsection{Impact on Energy Performance}
\begin{figure}[h!]
    \centering
    \vspace{-3mm}
    \includegraphics[width=\linewidth]{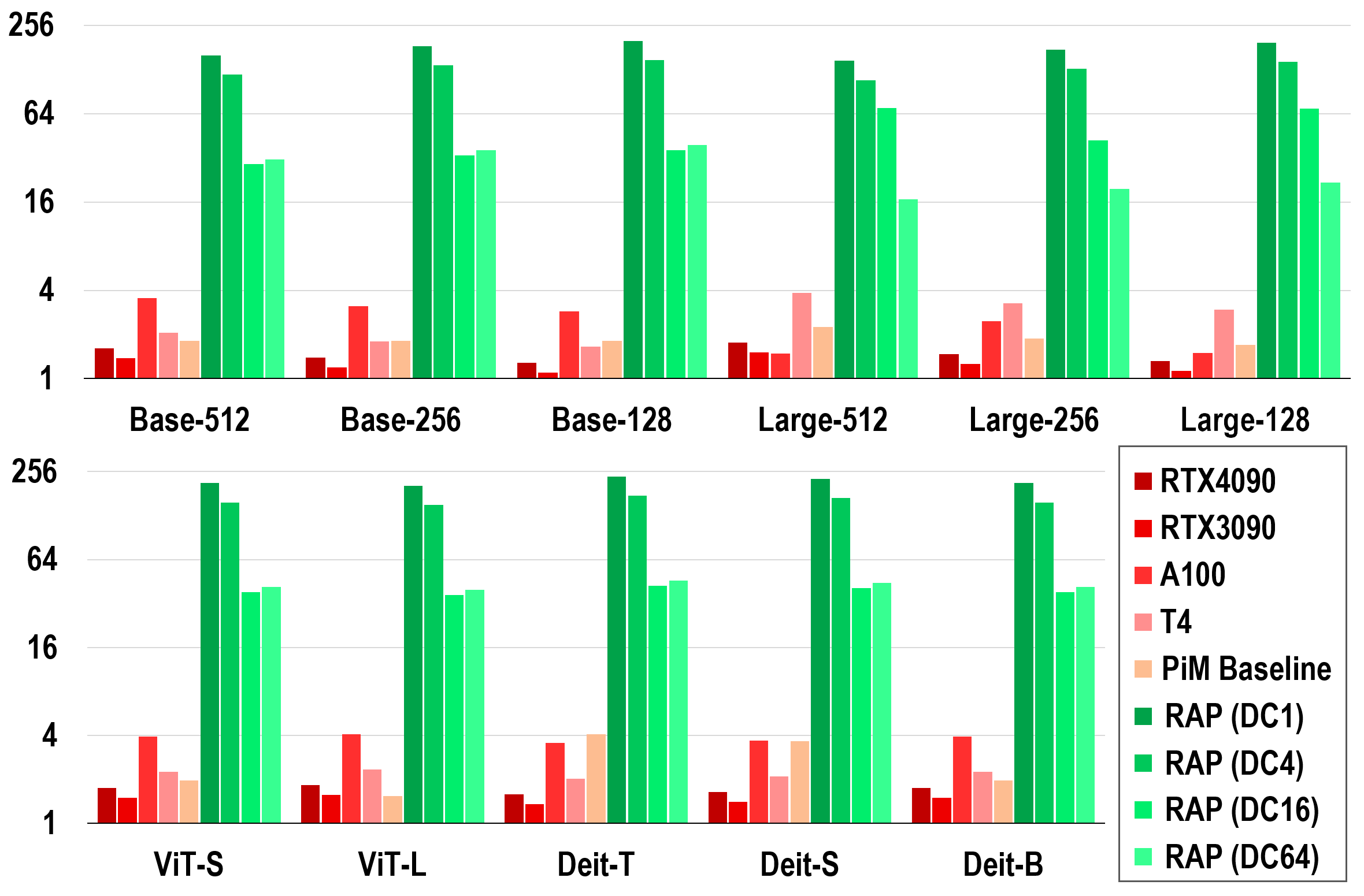}
    \vspace{-7.5mm}
    \caption{Comparison of normalized token/Joule performance.}
    \vspace{-3mm}
    \label{fig:comparison_Joule_per_Layer}
\end{figure}
Fig.~\ref{fig:comparison_Joule_per_Layer} illustrates the energy efficiency of our FLARE architecture compared to various SOTA GPUs and a PiM baseline across diverse NLP and vision tasks. The results highlight that our architecture achieves outstanding energy efficiency, positioning it as a practical solution for energy-constrained applications. Notably, without our on-device end-to-end kernel fusion technique, the reliance on FPUs or the substantial out-of-PiM tensor traffic significantly undermines the baseline PiM devices' inherent efficiency. 

%Our experimental results highlight the efficacy of the FLARE architecture in terms of latency and energy efficiency. The design exploration underscores the importance of customizing inner-array configurations and parallelism levels to achieve optimal performance. The proposed eMSB-Q and VDR-Softmax techniques effectively maintain inference accuracy while eliminating FP operations, thus enhancing computational efficiency. Additionally, the BitSift-GEMV technique significantly boosts GEMV operations by leveraging bitwise sparsity. Overall, ourarchitecture offers a robust and scalable solution for accelerating transformer models, providing substantial improvements in both latency and energy consumption.

\vspace{-1mm}
\section{Discussion - Related Works}
FLARE’s design offers an accurate, fast, and efficient foundation for accelerating self-attention layers in encoder models. Beyond its standalone capabilities, FLARE seamlessly integrates with optimization techniques to further enhance performance without hardware revisions.
For instance, FLARE complements FlashAttention~\cite{dao2022flashattention} by processing tiles independently in a FLARE PE, effectively scaling to long sequences or oversized models while preserving its core strengths. 
Also, FLARE aligns with Dynamic Attention Modulation techniques~\cite{moitra2024pivot}, dynamically allocating resources based on input complexity to reduce redundant computations and optimize energy efficiency.
Moreover, FLARE supports hot-expert routing~\cite{kim2024monde} in Mixture-of-Experts models, efficiently handling intensive computations while reducing tensor traffic. 
Finally, representation compression techniques~\cite{lan2019albert,jiao2019tinybert,sanh2019distilbert,5432202} align with FLARE’s processing to lower computational demands without sacrificing accuracy.

%In addition to the techniques proposed in this work, which significantly enhance the efficiency and performance of self-attention-layer acceleration, our FLARE architecture also opens avenues for integration with other optimization methods. 
%For instance, FLARE can complement Linformer~\cite{wang2020linformer}, which reduces the quadratic token-to-token complexity into a linear form using tiled attention, amplifying runtime improvements in energy efficiency and latency. Similarly, FLARE could synergize with dynamic attention modulation approaches like PIVOT~\cite{moitra2024pivot}, where resource allocation is guided by input complexity, using FLARE’s architecture to process critical paths with high efficiency.
%Furthermore, FLARE is well-suited for collaboration with traffic-compression techniques, such as hot-expert-selective Mixture-of-Experts (MoE) models~\cite{kim2024monde}, where computationally intensive “hot” experts are routed to FLARE for tensor traffic reduction. 

\vspace{-1mm}
\section{Conclusion}
This work presents FLARE, an AMS-PiM-based architecture designed to overcome the computational and hardware bottlenecks of transformer models. By introducing dequantization-free PTQ, integer-only nonlinear processing, and BitSift-GEMV for sparse GEMV acceleration, FLARE achieves robust energy efficiency, error resilience, and computational performance.
FLARE’s hybrid MRAM-SRAM design enables on-chip processing of self-attention layers while eliminating the need for high-ENOB ADCs and FPUs, reducing area and power overhead. Experimental results confirm FLARE's superiority over GPUs and PiM baselines in latency, energy efficiency, and scalability, making it a practical solution for deploying transformers in diverse environments.

\bibliographystyle{ACM-Reference-Format}

%\bibliography{references}


\end{document}